\documentclass{article}
\usepackage{ijcai24}
\newcommand{\citet}[1]{\citeauthor{#1}~\shortcite{#1}}
\newcommand{\citep}[1]{\cite{#1}}

\usepackage{times}
\usepackage{soul}
\usepackage{url}
\usepackage[hidelinks]{hyperref}
\usepackage[utf8]{inputenc}
\usepackage[small]{caption}
\usepackage{graphicx}
\usepackage{amsmath}
\usepackage{amsthm}
\usepackage{booktabs}
\usepackage{algorithm}
\usepackage{algorithmic}
\usepackage[switch]{lineno}
\usepackage{longtable}
\usepackage{minitoc}

\usepackage{amssymb}
\usepackage{tikz}
\usepackage{capt-of}
\usepackage{svg}
\usepackage{mathtools}
\usepackage{subcaption}
\usepackage{adjustbox}
\usepackage{diagbox}
\usepackage{xcolor}

\newcommand{\DAG}{UDFS} 

\title{Scaling Up Unbiased Search-based Symbolic Regression}

\author {
Paul Kahlmeyer$^1$\and\,
Joachim Giesen$^1$\and
Michael Habeck$^{2}$\And
Henrik Voigt$^1$
\affiliations
$^1$Friedrich Schiller University Jena\\
$^2$University Hospital Jena\\
\emails
\{paul.kahlmeyer, joachim.giesen, michael.habeck, henrik.voigt\}@uni-jena.de
}

\begin{document}

\maketitle

\begin{abstract}
In a regression task, a function is learned from labeled data to predict the labels at new data points. The goal is to achieve small prediction errors. In symbolic regression, the goal is more ambitious, namely, to learn an \emph{interpretable} function that makes small prediction errors. This additional goal largely rules out the standard approach used in regression, that is, reducing the learning problem to learning parameters of an expansion of basis functions by optimization. Instead, symbolic regression methods \emph{search} for a good solution in a space of symbolic expressions. To cope with the typically vast search space, most symbolic regression methods make implicit, or sometimes even explicit, assumptions about its structure. Here, we argue that the only obvious structure of the search space is that it contains small expressions, that is, expressions that can be decomposed into a few subexpressions. We show that systematically searching spaces of small expressions finds solutions that are more accurate and more robust against noise than those obtained by state-of-the-art symbolic regression methods. In particular, systematic search outperforms state-of-the-art symbolic regressors in terms of its ability to recover the true underlying symbolic expressions on established benchmark data sets\footnote{\href{https://github.com/kahlmeyer94/DAG\_search}{\textcolor{blue}{\texttt{https://github.com/kahlmeyer94/DAG\_search}}}}.
\end{abstract}

\section{Introduction}
\label{introduction}

Given a training set of labeled data, the goal in regression is to find a model that generalizes well beyond the training data.
At its core, regression problems are search problems on some space of functions that map data points to labels.
Typically, the function space is structured along two dimensions, a space of \emph{structural frames} and a set of \emph{parameters}.
Traditionally, the search space of structural frames is kept rather simple, as a set of linear combinations of some basis functions.
For a given regression problem, by considering only one frame, that is, traditionally the number and form of basis functions, the search problem can be cast as the optimization problem to find the parameters that give the best fit on the training data. 
Therefore, the focus shifts to the parameter space. Restricting the parameter space by adding regularization terms to the optimization problem can provably improve the ability to generalize~\cite{Hoerl70,Tibshirani96}.

\begin{figure}[t]
    \centering
    \includegraphics[width = \columnwidth]{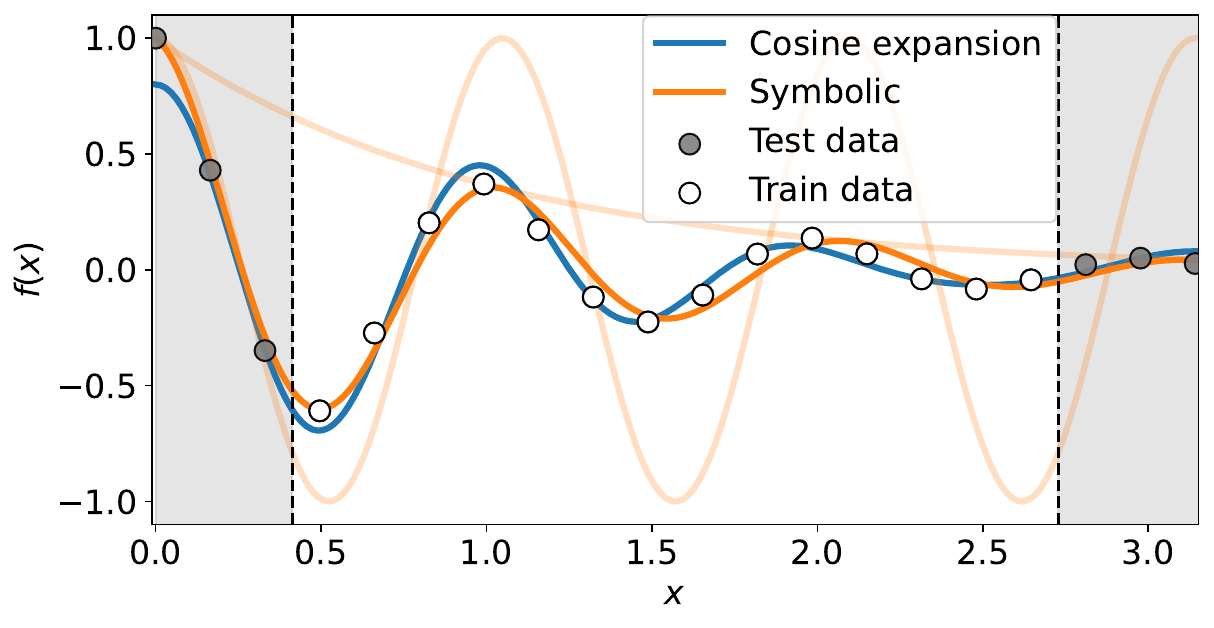}
    \caption{Interpretability goes beyond generalization. Data from a dampened pendulum are fitted by the symbolic regressor $e^{-x} \cos(6x)$ and the cosine expansion $0.06\cos(4x)+0.13\cos(5x)+0.30\cos(6x)+0.23\cos(7x)+0.08\cos(8x)$. While both models generalize well outside the training data, the symbolic regressor can be interpreted as the multiplicative composition of an exponential decay and an oscillation (light orange). The cosine expansion lacks such an interpretation.}
    \label{fig: oscillation}
\end{figure}

Generalization is not the only goal of symbolic regression.
The second goal is interpretability.
Interpretability, facilitating fundamental insights by inspecting and interpreting the symbolic expressions, is the reason why symbolic regression has found applications in almost all areas of the natural sciences~\cite{keren23,liu21,liu22}, social sciences~\cite{aryadoust15,truscott11}, and engineering~\cite{CAN2011447,quade16,WangWR19,dsr4_petersen21}.
We illustrate this point using the example shown in Figure~\ref{fig: oscillation}.
Data points that are sampled from a dampened pendulum are fitted by two models. 
The first model is a linear combination of cosine basis functions (blue) and the second model is a symbolic regressor (orange).
Both models generalize well beyond the training data and could be used for prediction.
From the symbolic form of the polynomial, however, we do not gain further insights. 
The symbolic regressor provides such insights: it is composed of an exponential decay and an oscillation that are coupled through multiplication, meaning that the oscillation is dampened by the decay. 

Because of the second goal, interpretability, symbolic regression algorithms are evaluated differently than standard regression algorithms. 
Good prediction performance on test data is still important, but for symbolic regression algorithms the ability to recover known ground-truth formulas, up to symbolic equivalence, has become an accepted validation measure.
Complete symbolic recovery, however, is a very strict quality measure.
Therefore, we propose also a relaxed measure, namely, the successful recovery of subexpressions.
For instance, recovering only the dampening or only the oscillation term of the dampened pendulum, still provides valuable insights.

Interpretability of symbolic regressors comes at the price of a much larger search space of structural frames, which consists of all mathematical expressions that are specified in a formal language, where constants are not instantiated but represented by a placeholder symbol.
State-of-the-art symbolic regressors tackle the search problem by various techniques such as genetic programming, reinforcement learning, or transformers that all make some implicit assumptions to reduce the effective size of the search space. 
In this work, we take a step back and explore a basic, unbiased and thus assumption-free search.
It turns out, that this basic approach fares well in comparison to current state-of-the-art regressors on small instances of established benchmark data sets.
It does not, however, scale to moderately large expressions.
For scaling up the basic search, we combine it with a variable augmentation approach that aims at identifying subexpressions in the target expression that can be eliminated from the search process.
Variable augmentation itself constitutes a search problem for subexpressions, that can be addressed by the same technique as the overall symbolic regression task.
Our experimental results on the standard benchmark data sets show that the combination of unbiased search and variable augmentation improves the state of the art in terms of \emph{accuracy}, that is, its ability to recover known ground truth formulas, but also in terms of \emph{robustness}, that is, its ability to cope with noise. 

\section{Related Work}

The core problem in symbolic regression is searching the space of structural frames.
For a given frame, the parameters are mostly estimated by minimizing a loss function on hold-out data.
The space of frames is usually given in the form of expression trees for expressions from a given formal language.
Symbolic regression algorithms differ in the way they search the space of expression trees. 

\paragraph{Genetic Programming.}

The majority of symbolic regression algorithms follow the \emph{genetic programming} paradigm~\cite{koza1994genetic,Holland:1975}. The paradigm has been implemented in the seminal Eureqa system by~\cite{eureqa_lipson09} and in gplearn by \cite{gplearn_stephens16}. More recent implementations include~\cite{eplex_lacava16,kommenda2020parameter,virgolin2021improving}. The basic idea is to create a population of expression trees, to turn the trees into symbolic regressors by estimating the parameters, and to recombine the expression trees for the best performing regressors into a new population of expression trees by (ex)changing subtrees. Here, the assumption is, that expression trees for symbolic regressors that perform well contain at least parts of the target expression. Therefore, recombining the best performing expression trees shall keep these parts within the population.

\paragraph{Bayesian Inference.}

A different approach is to use Bayesian inference for symbolic regression.
In the Bayesian inference approach, MCMC (Markov Chain Monte Carlo) is used for sampling expressions from a posterior distribution. 
Implementations of the Bayesian approach differ in the choice of prior distribution.
\cite{jin2020bayesian} use a hand-designed prior distribution on expression trees, whereas \cite{bayesianscientist_guimera20} compile a prior distribution from a corpus of 4,080 mathematical expressions that have been extracted from Wikipedia articles on problems from physics and the social sciences.
Both implementations define the likelihood in terms of model fit.
Here, the assumption is, that, by the choice of prior and likelihood, expression trees that are similar to well performing trees have a higher posterior probability to be sampled.

\paragraph{Neural Networks.}

\cite{dsr_petersen21,dsr2_petersen21} train a recurrent neural network for sampling expression trees token-by-token in preorder. Here, the assumption is similar to the assumption underlying the Bayesian approach, namely that the loss used during reinforcement learning shifts the token-generating distribution toward sequences of good performing expressions.
The work by \cite{transformer_kamienny22} also falls into this category. Here, a transformer is trained in an end-to-end fashion on a large data set of regression problems to translate regression tasks, given as a sequence of input output pairs, into a sequence of tokens for an expression tree in preorder.
At inference, for a given regression problem, beam search is used to return preorders with the highest likelihood.
Here, the assumption is that the data set for pretraining the transformer covers the search space well, because the probability distribution favors token sequences that are similar to well-fitting sequences seen during training.

\paragraph{Ensemble.}

Recently, \cite{dsr3_petersen22} have combined different symbolic regressors based on genetic programming, reinforcement learning, and transformers together with problem simplification into a unified symbolic regressor. The ensemble approach is more robust with respect to the assumptions for its constituent approaches, that is, it can cope better with some of the assumptions not met. Essentially, however, it adheres to the same assumptions.

\paragraph{Systematic Search.}

The assumptions made for the different symbolic regression approaches are difficult to check.
Therefore, the idea of an unbiased, more or less assumption-free search is appealing.
So far, however, implementations of an unbiased search have been limited to restricted search spaces that can be searched exhaustively, namely, to rational low-degree polynomials with linear and nonlinear terms \cite{rationals_kammerer20}, and to univariate regression problems up to a certain depth \cite{esr_bartlett23}.
The AIFeynman project \cite{feynmanAI_udrescu20,feynmanAI2_udrescu20} uses a set of neural-network-based statistical property tests to scale an unbiased search to larger search spaces.
Statistically significant properties, for example, additive or multiplicative separability, are used to decompose the search space into smaller parts, which are addressed by a brute-force search.
The search, however, is limited by the set of available property tests. 
Our approach is similar in spirit to AIFeynman, but different in implementation.
At the core of our approach is a succinct representation of symbolic expressions, namely expression DAGs that are, as we will show, well suited for a systematic unbiased, search-based approach.

\section{DAGs as Structural Frames}

The search space of structural frames in symbolic regression consists of mathematical expressions that are defined by a formal language.
We provide the grammar for the formal language of mathematical expressions that we are using here in the supplemental material.
The grammar, however, does not completely specify the search space.
There are two issues that affect any search-based approach to symbolic regression:
First, the grammar specifies the syntax of valid expressions, but does not fix the representation for its conforming expressions. Second, by the rules of arithmetic, there are syntactically different expressions that are semantically equivalent, that is, they specify exactly the same function.
Therefore, it is important to find a representation for the expressions that structures the search space so that the number of functions that can be covered for a given size constraint on the search space is maximized.
The most direct representation of an expression is in the form of a string of tokens, but the most commonly used representation in symbolic regression is an expression tree. 
We illustrate the two issues on the example of the function 
\[
f(x)=x^4 + x^3\,,
\]
which has an expression tree with five operator nodes (addition, multiplication, and three times squaring). 
The same function, however, has another expression as $x^2(x^2 + x)$ with an expression tree that has only four operator nodes (addition, multiplication, and two times squaring).
Therefore, it is often suffcient to exhaustively search the space of small expression trees.

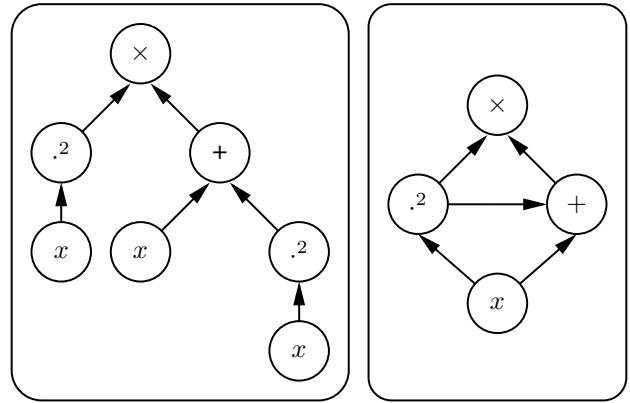
\begin{figure}[h!]
     \centering
     \tikzset{every picture/.style={line width=0.75pt}} 

\begin{tikzpicture}[x=0.75pt,y=0.75pt,yscale=-1,xscale=1]

\draw    (215,111) -- (278,111.03) ;
\draw [shift={(280,111.03)}, rotate = 180.03] [fill={rgb, 255:red, 0; green, 0; blue, 0 }  ][line width=0.08]  [draw opacity=0] (12,-3) -- (0,0) -- (12,3) -- cycle    ;
\draw    (295,111) -- (261.41,77.41) ;
\draw [shift={(260,76)}, rotate = 45] [fill={rgb, 255:red, 0; green, 0; blue, 0 }  ][line width=0.08]  [draw opacity=0] (12,-3) -- (0,0) -- (12,3) -- cycle    ;
\draw    (215,111) -- (248.59,77.41) ;
\draw [shift={(250,76)}, rotate = 135] [fill={rgb, 255:red, 0; green, 0; blue, 0 }  ][line width=0.08]  [draw opacity=0] (12,-3) -- (0,0) -- (12,3) -- cycle    ;
\draw    (255,161) -- (216.51,127.32) ;
\draw [shift={(215,126)}, rotate = 41.19] [fill={rgb, 255:red, 0; green, 0; blue, 0 }  ][line width=0.08]  [draw opacity=0] (12,-3) -- (0,0) -- (12,3) -- cycle    ;
\draw    (255,161) -- (293.53,127.32) ;
\draw [shift={(295.03,126)}, rotate = 138.84] [fill={rgb, 255:red, 0; green, 0; blue, 0 }  ][line width=0.08]  [draw opacity=0] (12,-3) -- (0,0) -- (12,3) -- cycle    ;
\draw  [fill={rgb, 255:red, 255; green, 255; blue, 255 }  ,fill opacity=1 ] (240,161) .. controls (240,152.72) and (246.72,146) .. (255,146) .. controls (263.28,146) and (270,152.72) .. (270,161) .. controls (270,169.28) and (263.28,176) .. (255,176) .. controls (246.72,176) and (240,169.28) .. (240,161) -- cycle ;
\draw  [fill={rgb, 255:red, 255; green, 255; blue, 255 }  ,fill opacity=1 ] (280,111.03) .. controls (279.98,102.75) and (286.68,96.02) .. (294.97,96) .. controls (303.25,95.98) and (309.98,102.68) .. (310,110.97) .. controls (310.02,119.25) and (303.32,125.98) .. (295.03,126) .. controls (286.75,126.02) and (280.02,119.32) .. (280,111.03) -- cycle ;
\draw  [fill={rgb, 255:red, 255; green, 255; blue, 255 }  ,fill opacity=1 ] (200,111) .. controls (200,102.72) and (206.72,96) .. (215,96) .. controls (223.28,96) and (230,102.72) .. (230,111) .. controls (230,119.28) and (223.28,126) .. (215,126) .. controls (206.72,126) and (200,119.28) .. (200,111) -- cycle ;
\draw  [fill={rgb, 255:red, 255; green, 255; blue, 255 }  ,fill opacity=1 ] (240,61) .. controls (240,52.72) and (246.72,46) .. (255,46) .. controls (263.28,46) and (270,52.72) .. (270,61) .. controls (270,69.28) and (263.28,76) .. (255,76) .. controls (246.72,76) and (240,69.28) .. (240,61) -- cycle ;

\draw    (155,135) -- (121.41,101.41) ;
\draw [shift={(120,100)}, rotate = 45] [fill={rgb, 255:red, 0; green, 0; blue, 0 }  ][line width=0.08]  [draw opacity=0] (12,-3) -- (0,0) -- (12,3) -- cycle    ;
\draw    (75,135) -- (108.59,101.41) ;
\draw [shift={(110,100)}, rotate = 135] [fill={rgb, 255:red, 0; green, 0; blue, 0 }  ][line width=0.08]  [draw opacity=0] (12,-3) -- (0,0) -- (12,3) -- cycle    ;
\draw    (115,85) -- (81.41,51.41) ;
\draw [shift={(80,50)}, rotate = 45] [fill={rgb, 255:red, 0; green, 0; blue, 0 }  ][line width=0.08]  [draw opacity=0] (12,-3) -- (0,0) -- (12,3) -- cycle    ;
\draw    (35,85) -- (68.59,51.41) ;
\draw [shift={(70,50)}, rotate = 135] [fill={rgb, 255:red, 0; green, 0; blue, 0 }  ][line width=0.08]  [draw opacity=0] (12,-3) -- (0,0) -- (12,3) -- cycle    ;
\draw  [fill={rgb, 255:red, 255; green, 255; blue, 255 }  ,fill opacity=1 ] (60,35) .. controls (60,26.72) and (66.72,20) .. (75,20) .. controls (83.28,20) and (90,26.72) .. (90,35) .. controls (90,43.28) and (83.28,50) .. (75,50) .. controls (66.72,50) and (60,43.28) .. (60,35) -- cycle ;
\draw  [fill={rgb, 255:red, 255; green, 255; blue, 255 }  ,fill opacity=1 ] (100,85) .. controls (100,76.72) and (106.72,70) .. (115,70) .. controls (123.28,70) and (130,76.72) .. (130,85) .. controls (130,93.28) and (123.28,100) .. (115,100) .. controls (106.72,100) and (100,93.28) .. (100,85) -- cycle ;
\draw   (10,25.11) .. controls (10,16.77) and (16.77,10) .. (25.11,10) -- (164.89,10) .. controls (173.23,10) and (180,16.77) .. (180,25.11) -- (180,194.89) .. controls (180,203.23) and (173.23,210) .. (164.89,210) -- (25.11,210) .. controls (16.77,210) and (10,203.23) .. (10,194.89) -- cycle ;
\draw   (190,21.56) .. controls (190,15.17) and (195.17,10) .. (201.56,10) -- (308.44,10) .. controls (314.83,10) and (320,15.17) .. (320,21.56) -- (320,198.44) .. controls (320,204.83) and (314.83,210) .. (308.44,210) -- (201.56,210) .. controls (195.17,210) and (190,204.83) .. (190,198.44) -- cycle ;
\draw  [fill={rgb, 255:red, 255; green, 255; blue, 255 }  ,fill opacity=1 ] (60,135) .. controls (60,126.72) and (66.72,120) .. (75,120) .. controls (83.28,120) and (90,126.72) .. (90,135) .. controls (90,143.28) and (83.28,150) .. (75,150) .. controls (66.72,150) and (60,143.28) .. (60,135) -- cycle ;
\draw  [fill={rgb, 255:red, 255; green, 255; blue, 255 }  ,fill opacity=1 ] (140,135) .. controls (140,126.72) and (146.72,120) .. (155,120) .. controls (163.28,120) and (170,126.72) .. (170,135) .. controls (170,143.28) and (163.28,150) .. (155,150) .. controls (146.72,150) and (140,143.28) .. (140,135) -- cycle ;
\draw    (155,185) -- (155,152) ;
\draw [shift={(155,150)}, rotate = 90] [fill={rgb, 255:red, 0; green, 0; blue, 0 }  ][line width=0.08]  [draw opacity=0] (12,-3) -- (0,0) -- (12,3) -- cycle    ;
\draw  [fill={rgb, 255:red, 255; green, 255; blue, 255 }  ,fill opacity=1 ] (140,185) .. controls (140,176.72) and (146.72,170) .. (155,170) .. controls (163.28,170) and (170,176.72) .. (170,185) .. controls (170,193.28) and (163.28,200) .. (155,200) .. controls (146.72,200) and (140,193.28) .. (140,185) -- cycle ;
\draw  [fill={rgb, 255:red, 255; green, 255; blue, 255 }  ,fill opacity=1 ] (20,85) .. controls (20,76.72) and (26.72,70) .. (35,70) .. controls (43.28,70) and (50,76.72) .. (50,85) .. controls (50,93.28) and (43.28,100) .. (35,100) .. controls (26.72,100) and (20,93.28) .. (20,85) -- cycle ;
\draw    (35,130) -- (35,102) ;
\draw [shift={(35,100)}, rotate = 90] [fill={rgb, 255:red, 0; green, 0; blue, 0 }  ][line width=0.08]  [draw opacity=0] (12,-3) -- (0,0) -- (12,3) -- cycle    ;
\draw  [fill={rgb, 255:red, 255; green, 255; blue, 255 }  ,fill opacity=1 ] (20,135) .. controls (20,126.72) and (26.72,120) .. (35,120) .. controls (43.28,120) and (50,126.72) .. (50,135) .. controls (50,143.28) and (43.28,150) .. (35,150) .. controls (26.72,150) and (20,143.28) .. (20,135) -- cycle ;

\draw (75,35) node    {$\times $};
\draw (115,85) node   [align=left] {+};
\draw (75,135) node    {$x$};
\draw (155,185) node    {$x$};
\draw (35,85) node    {$\cdot ^{2}$};
\draw (35,135) node    {$x$};
\draw (155,135) node    {$\cdot ^{2}$};
\draw (255,161) node    {$x$};
\draw (295,111) node  [rotate=-359.88]  {$+$};
\draw (255,61) node    {$\times $};
\draw (215,111) node    {$\cdot ^{2}$};

\end{tikzpicture}
     \caption{Representation of the expression $x^2(x^2+x)$ by an expression tree (left) and by an expression DAG (right).}
    \label{fig:representation_examples}
\end{figure}

Moreover, it has been pointed out already by \cite{graphs_lipson07} and is also well known in compiler construction \cite{compiler1986} that expression DAGs (directed acyclic graphs), where common subexpressions have been eliminated \cite{CSE}, are an even more favorable because more succinct representation.
The expression DAG for the expression $x^2(x^2 + x)$, that factors out the common subexpression $x^2$, has only three operator nodes (addition, multiplication, and squaring).
The difference in size becomes more pronounced when we also consider the leafs, that either store variables or constants, of the expression trees and expression DAGs, respectively. 
We illustrate the difference between the expression tree and expression DAG representation in Figure~\ref{fig:representation_examples}.

\paragraph{Expression DAGs.}

Here, we describe the expression DAGs that we use for arithmetic expressions in more detail.
We distinguish four types of nodes: variable nodes, parameter nodes, intermediary nodes, and output nodes. 
Variable and parameter nodes are \emph{input} nodes. 
An $n$-variate function has $n$ input nodes for the variables, that is, it is defined by the regression task at hand, whereas the number of input nodes for the parameters is not fixed, but part of the structural search space.
Both, the intermediary and the output nodes together are \emph{operator} nodes.
The number of output nodes is also specified by the given regression task, whereas the number of intermediary nodes is not.
Note that it can pay off to encode a function with several components, that is, several output nodes, into a single expression DAG, because the components can share common subexpressions.
This happens frequently for systems of ordinary differential equations~\cite{Strogatz00}.
The DAGs are oriented from the input to the output nodes, that is, the input nodes are the roots of the DAGs and the output nodes are its leafs.
Only the intermediary nodes have incoming and outgoing edges.
We distinguish two types of operator nodes, namely unary and binary operator nodes.
In symbolic regression, typically, the binary operators 
\[
+,\; -,\; \times, \;\textrm{ and }\; \div, 
\]
and the unary operators 
\[
-,\; {}^{-1},\; \sin,\;  \cos,\; \log,\; \exp,\; {}^2, \;\textrm{ and }\; \sqrt{\phantom{x}}
\]
are supported. 
Examples of expression DAGs are shown in Figures~\ref{fig:representation_examples} and~\ref{fig:frames}. 

\section{Searching the Space of Expression DAGs}
\label{sec:DAGSearch}

As we have pointed out already, the number of variable nodes and the number of output nodes are specified by the given regression problem.
It remains to parameterize the space of expression DAG by the number $p$ of parameter nodes and by the number $i$ of intermediary nodes.
That is, the regression problem together with a tuple $(p,i)$ defines the search space.
Here, we always use $p=1$, that is, only one parameter node.
Different parameters can be expressed as functions of a single parameter, for instance $2(x^2+1)$ can be expressed as $(1+1)(x^2+1)$ or as $2x^2+2$, which only need one parameter.
More details are provided in the supplemental material.
We use the term DAG \emph{skeleton} for DAGs with unlabeled operator nodes, and the term DAG \emph{frame} for DAGs with labeled operator nodes.
Both, DAG skeletons and DAG frames, only have constant placeholders for the input nodes.

\subsection{Unbiased Search of Expression DAGs}
\label{sec:unbiasedDAGSearch}

Our randomized search of the search space of expression DAGs given by $(1,i)$ is unbiased, but not exhaustive. 
An exhaustive search would not scale and, as it turns out (see section~\ref{sec:experiments}), is often not necessary. 
The search procedure has two phases:
In the first phase, we construct a DAG skeleton, that is, a DAG without labeling of the operator nodes. 
In the second phase, we search over all DAG frames for a skeleton, by considering all operator node labelings from the sets of unary and binary operator symbols.

\paragraph{Sampling DAG Skeletons.}

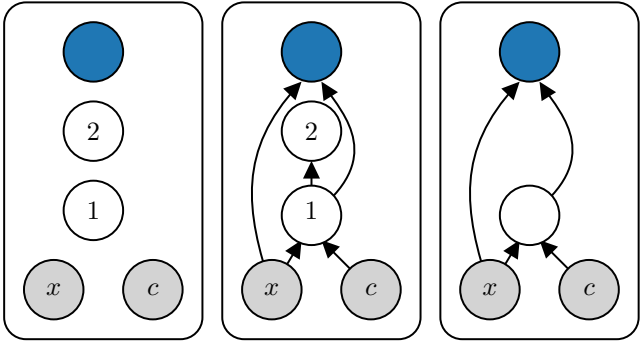
\begin{figure}[h!]
     \centering
     \tikzset{every picture/.style={line width=0.75pt}} 

\begin{tikzpicture}[x=0.75pt,y=0.75pt,yscale=-1,xscale=1]

\draw    (305,155) -- (282.12,132.12) ;
\draw [shift={(280,130)}, rotate = 45] [fill={rgb, 255:red, 0; green, 0; blue, 0 }  ][line width=0.08]  [draw opacity=0] (8.93,-4.29) -- (0,0) -- (8.93,4.29) -- cycle    ;
\draw    (255,155) -- (268.46,132.57) ;
\draw [shift={(270,130)}, rotate = 120.96] [fill={rgb, 255:red, 0; green, 0; blue, 0 }  ][line width=0.08]  [draw opacity=0] (8.93,-4.29) -- (0,0) -- (8.93,4.29) -- cycle    ;
\draw    (255,155) .. controls (241.28,114.33) and (238.12,84.7) .. (268.12,52.01) ;
\draw [shift={(270,50)}, rotate = 133.69] [fill={rgb, 255:red, 0; green, 0; blue, 0 }  ][line width=0.08]  [draw opacity=0] (8.93,-4.29) -- (0,0) -- (8.93,4.29) -- cycle    ;
\draw    (275,117.5) .. controls (312.8,89.15) and (291.66,67.05) .. (281.63,52.47) ;
\draw [shift={(280,50)}, rotate = 58.17] [fill={rgb, 255:red, 0; green, 0; blue, 0 }  ][line width=0.08]  [draw opacity=0] (8.93,-4.29) -- (0,0) -- (8.93,4.29) -- cycle    ;
\draw  [fill={rgb, 255:red, 211; green, 211; blue, 211 }  ,fill opacity=1 ] (240,155) .. controls (240,146.72) and (246.72,140) .. (255,140) .. controls (263.28,140) and (270,146.72) .. (270,155) .. controls (270,163.28) and (263.28,170) .. (255,170) .. controls (246.72,170) and (240,163.28) .. (240,155) -- cycle ;
\draw  [fill={rgb, 255:red, 255; green, 255; blue, 255 }  ,fill opacity=1 ] (260,117.5) .. controls (260,109.22) and (266.72,102.5) .. (275,102.5) .. controls (283.28,102.5) and (290,109.22) .. (290,117.5) .. controls (290,125.78) and (283.28,132.5) .. (275,132.5) .. controls (266.72,132.5) and (260,125.78) .. (260,117.5) -- cycle ;
\draw  [fill={rgb, 255:red, 31; green, 119; blue, 180 }  ,fill opacity=1 ] (260,35) .. controls (260,26.72) and (266.72,20) .. (275,20) .. controls (283.28,20) and (290,26.72) .. (290,35) .. controls (290,43.28) and (283.28,50) .. (275,50) .. controls (266.72,50) and (260,43.28) .. (260,35) -- cycle ;
\draw  [fill={rgb, 255:red, 211; green, 211; blue, 211 }  ,fill opacity=1 ] (290,155) .. controls (290,146.72) and (296.72,140) .. (305,140) .. controls (313.28,140) and (320,146.72) .. (320,155) .. controls (320,163.28) and (313.28,170) .. (305,170) .. controls (296.72,170) and (290,163.28) .. (290,155) -- cycle ;
\draw    (165,115) -- (165,93) ;
\draw [shift={(165,90)}, rotate = 90] [fill={rgb, 255:red, 0; green, 0; blue, 0 }  ][line width=0.08]  [draw opacity=0] (8.93,-4.29) -- (0,0) -- (8.93,4.29) -- cycle    ;
\draw    (195,155) -- (172.12,132.12) ;
\draw [shift={(170,130)}, rotate = 45] [fill={rgb, 255:red, 0; green, 0; blue, 0 }  ][line width=0.08]  [draw opacity=0] (8.93,-4.29) -- (0,0) -- (8.93,4.29) -- cycle    ;
\draw    (145,155) -- (158.46,132.57) ;
\draw [shift={(160,130)}, rotate = 120.96] [fill={rgb, 255:red, 0; green, 0; blue, 0 }  ][line width=0.08]  [draw opacity=0] (8.93,-4.29) -- (0,0) -- (8.93,4.29) -- cycle    ;
\draw    (145,155) .. controls (131.28,114.33) and (128.12,84.7) .. (158.12,52.01) ;
\draw [shift={(160,50)}, rotate = 133.69] [fill={rgb, 255:red, 0; green, 0; blue, 0 }  ][line width=0.08]  [draw opacity=0] (8.93,-4.29) -- (0,0) -- (8.93,4.29) -- cycle    ;
\draw    (165,117.5) .. controls (202.8,89.15) and (181.66,67.05) .. (171.63,52.47) ;
\draw [shift={(170,50)}, rotate = 58.17] [fill={rgb, 255:red, 0; green, 0; blue, 0 }  ][line width=0.08]  [draw opacity=0] (8.93,-4.29) -- (0,0) -- (8.93,4.29) -- cycle    ;
\draw  [fill={rgb, 255:red, 211; green, 211; blue, 211 }  ,fill opacity=1 ] (130,155) .. controls (130,146.72) and (136.72,140) .. (145,140) .. controls (153.28,140) and (160,146.72) .. (160,155) .. controls (160,163.28) and (153.28,170) .. (145,170) .. controls (136.72,170) and (130,163.28) .. (130,155) -- cycle ;
\draw  [fill={rgb, 255:red, 255; green, 255; blue, 255 }  ,fill opacity=1 ] (150,117.5) .. controls (150,109.22) and (156.72,102.5) .. (165,102.5) .. controls (173.28,102.5) and (180,109.22) .. (180,117.5) .. controls (180,125.78) and (173.28,132.5) .. (165,132.5) .. controls (156.72,132.5) and (150,125.78) .. (150,117.5) -- cycle ;
\draw  [fill={rgb, 255:red, 255; green, 255; blue, 255 }  ,fill opacity=1 ] (150,75) .. controls (150,66.72) and (156.72,60) .. (165,60) .. controls (173.28,60) and (180,66.72) .. (180,75) .. controls (180,83.28) and (173.28,90) .. (165,90) .. controls (156.72,90) and (150,83.28) .. (150,75) -- cycle ;
\draw  [fill={rgb, 255:red, 31; green, 119; blue, 180 }  ,fill opacity=1 ] (150,35) .. controls (150,26.72) and (156.72,20) .. (165,20) .. controls (173.28,20) and (180,26.72) .. (180,35) .. controls (180,43.28) and (173.28,50) .. (165,50) .. controls (156.72,50) and (150,43.28) .. (150,35) -- cycle ;
\draw  [fill={rgb, 255:red, 211; green, 211; blue, 211 }  ,fill opacity=1 ] (180,155) .. controls (180,146.72) and (186.72,140) .. (195,140) .. controls (203.28,140) and (210,146.72) .. (210,155) .. controls (210,163.28) and (203.28,170) .. (195,170) .. controls (186.72,170) and (180,163.28) .. (180,155) -- cycle ;
\draw  [fill={rgb, 255:red, 211; green, 211; blue, 211 }  ,fill opacity=1 ] (20,155) .. controls (20,146.72) and (26.72,140) .. (35,140) .. controls (43.28,140) and (50,146.72) .. (50,155) .. controls (50,163.28) and (43.28,170) .. (35,170) .. controls (26.72,170) and (20,163.28) .. (20,155) -- cycle ;
\draw  [fill={rgb, 255:red, 255; green, 255; blue, 255 }  ,fill opacity=1 ] (40,115) .. controls (40,106.72) and (46.72,100) .. (55,100) .. controls (63.28,100) and (70,106.72) .. (70,115) .. controls (70,123.28) and (63.28,130) .. (55,130) .. controls (46.72,130) and (40,123.28) .. (40,115) -- cycle ;
\draw  [fill={rgb, 255:red, 255; green, 255; blue, 255 }  ,fill opacity=1 ] (40,75) .. controls (40,66.72) and (46.72,60) .. (55,60) .. controls (63.28,60) and (70,66.72) .. (70,75) .. controls (70,83.28) and (63.28,90) .. (55,90) .. controls (46.72,90) and (40,83.28) .. (40,75) -- cycle ;
\draw  [fill={rgb, 255:red, 31; green, 119; blue, 180 }  ,fill opacity=1 ] (40,35) .. controls (40,26.72) and (46.72,20) .. (55,20) .. controls (63.28,20) and (70,26.72) .. (70,35) .. controls (70,43.28) and (63.28,50) .. (55,50) .. controls (46.72,50) and (40,43.28) .. (40,35) -- cycle ;
\draw  [fill={rgb, 255:red, 211; green, 211; blue, 211 }  ,fill opacity=1 ] (70,155) .. controls (70,146.72) and (76.72,140) .. (85,140) .. controls (93.28,140) and (100,146.72) .. (100,155) .. controls (100,163.28) and (93.28,170) .. (85,170) .. controls (76.72,170) and (70,163.28) .. (70,155) -- cycle ;
\draw   (10,21) .. controls (10,14.92) and (14.92,10) .. (21,10) -- (99,10) .. controls (105.08,10) and (110,14.92) .. (110,21) -- (110,169) .. controls (110,175.08) and (105.08,180) .. (99,180) -- (21,180) .. controls (14.92,180) and (10,175.08) .. (10,169) -- cycle ;
\draw   (120,21) .. controls (120,14.92) and (124.92,10) .. (131,10) -- (209,10) .. controls (215.08,10) and (220,14.92) .. (220,21) -- (220,169) .. controls (220,175.08) and (215.08,180) .. (209,180) -- (131,180) .. controls (124.92,180) and (120,175.08) .. (120,169) -- cycle ;
\draw   (230,21) .. controls (230,14.92) and (234.92,10) .. (241,10) -- (319,10) .. controls (325.08,10) and (330,14.92) .. (330,21) -- (330,169) .. controls (330,175.08) and (325.08,180) .. (319,180) -- (241,180) .. controls (234.92,180) and (230,175.08) .. (230,169) -- cycle ;

\draw (255,155) node    {$x$};
\draw (305,155) node    {$c$};
\draw (165,115) node    {$1$};
\draw (165,75) node    {$2$};
\draw (145,155) node    {$x$};
\draw (195,155) node    {$c$};
\draw (55,115) node    {$1$};
\draw (55,75) node    {$2$};
\draw (35,155) node    {$x$};
\draw (85,155) node    {$c$};

\end{tikzpicture}
     \caption{DAG skeletons are generated by enumerating intermediary nodes (left), selecting predecessors according to the numbering (middle), and deleting intermediary nodes without connection to the output (right). The shown DAG covers expressions such as $x(x+c)$, $x + xc$ or $cx^2$.}
    \label{fig:frames}
\end{figure}

For sampling DAG skeletons, we number the intermediary nodes from $1$ to $i$ to ensure a topological order on the nodes.
\begin{enumerate}
    \item Unary output nodes sample its predecessor uniformly at random from all non-output nodes, and binary output nodes samples a pair of predecessors uniformly at random from all pairs of non-output nodes.
    \item Unary intermediary nodes sample their predecessor uniformly at random from all input nodes and all intermediary nodes with smaller number, and binary intermediary nodes sample a pair of predecessors uniformly at random from all pairs made up from input nodes and intermediary nodes with smaller number. 
\end{enumerate}
Finally, we recursively remove all intermediary nodes that have no successor, that is, no outgoing edge.
Note that input nodes have by definition no predecessors.
Sampling of a DAG skeleton is illustrated in Figure~\ref{fig:frames}.

\paragraph{Operator Node Labeling.}

For a DAG skeleton, we exhaustively search the space of all DAG frames, that is, we consider all combinations of operator node labelings from the set of unary and binary operator labels for unary and binary operator nodes respectively. 

\subsection{Scoring Expression DAG Frames}

An expression DAG frame $\Delta$ is not a regressor yet.
It remains to find values for the DAG's parameter node.
Here, we follow the classical approach and optimize the parameters with respect to the model fit on training data. 
Let $\Delta(x,\theta)$ be the function that results when the DAG's input nodes are instantiated by $x\in\mathbb{R}^n$, where $n$ is the number of input nodes, and its parameter nodes are instantiated by the parameter vector $\theta\in\mathbb{R}^p$, where $p$ is the number of parameter nodes in $\Delta$.
Given training data $(x_1,y_1),\ldots,(x_\ell,y_\ell)\in \mathbb{R}^{n\times m}$ for a regression problem with $n$ input and $m$ output variables, we compute the parameter values for the parameter nodes by minimizing the following square-loss
\[
\hat\theta = \textrm{arg} \underset{\theta\in\mathbb{R}^p}{\textrm{min}} \: \sum_{i=1}^\ell \big\| \Delta (x_i,\theta) - y_i\big\|^2 =: \textrm{arg} \underset{\theta\in\mathbb{R}^p}{\textrm{min}}\, L(\Delta,\theta).
\]
Every function $\Delta (\cdot,\hat\theta)$ is a regressor. 
Among all the regressors $\Delta (\cdot,\hat\theta)$ from the search space, that is, a search over $\Delta$, we choose one that fits the data best, that is, has a minimal loss $L(\Delta,\hat\theta)$.

\section{Variable Augmentation}

So far, we have argued that expression DAGs provide rather small search spaces for symbolic regression.
However, even these search spaces grow so fast that they can only support very small expressions.
Here, we describe our main contribution, namely, we introduce symbolic variable augmentations for simplifying symbolic regression problems. 

\subsection{Input Variable Augmentation}

We describe the basic idea on a simple regression problem with two input and one output variable.
Assume that we are given data that have been sampled from the function
\[
f(x_1, x_2) = \frac{x_1x_2^2 + x_1}{x_2}
\]
that has a corresponding expression DAG with six nodes, among them four operator nodes. 
Using a new variable $z(x_1,x_2) = x_1/x_2$, the function can be represented by the expression
\[
x_1x_2+z
\]
that has an expression with only five nodes, among them three input nodes. 
That is, it is possible to find a DAG frame for the function $f$ in a smaller search space, if we increase the number of input variables. 

Given a regression problem, we call any symbolic expression in the input variables a \emph{potential input variable augmentation}, where we only consider expressions without parameters ($p=0$). 
The challenge is to identify variable augmentations that lead to smaller expression DAGs. 
We address the challenge by a combination of searching expression DAGs and standard, that is, non-symbolic, regression.
The search of small expression DAGs is used to enumerate potential input variable augmentations, and a standard regressor is used to score the potential augmentations.

Given a regression problem, expression DAGs for potential variable augmentations are sampled using the unbiased expression DAG search from Section~\ref{sec:unbiasedDAGSearch}. 
Here, we sample only expression DAGs that feature a subset of the input variables nodes and no parameter nodes. 
Since the sampled expression DAGs $\Delta$ do not have parameter nodes, they each describe a unique univariate function $z_\Delta$ on the corresponding selected subset of the input variables. 
The functions $z_\Delta$ are then scored on training data $(x_1,y_1),\ldots,(x_\ell,y_\ell)\in \mathbb{R}^{n\times m}$ for a regression problem with $n$ input and $m$ output variables that are augmented as,
\[
(x_1,z_1,y_1),\ldots,(x_\ell,z_\ell,y_\ell)\in \mathbb{R}^{(n+1)\times m},
\]
where $z_i = z_\Delta (x_i|_\Delta) \in\mathbb{R}$ and $x_i|_\Delta$ is the projection of the $i$-th vector of input variables onto the input variables that appear in the expression DAG $\Delta$. 
For a given class $\mathcal{F}$ of standard regressors, for instance, polynomial regressors or a class of neural networks, we use a scoring function that is derived from the \emph{coefficient of determination} $R^2$~\cite{sewall_r2_1921}
\[
\min_{f\in\mathcal{F}} \: \frac{\sum_{i=1}^\ell \|  f(x_i,z_i) - y_i\|^2}{\sum_{i=1}^\ell \| y_i -\bar y\|^2} \,=:\, \min_{f\in\mathcal{F}} \: \big( 1 - R^2 (X,Y,f)\big)
,
\]
which is frequently used in symbolic regression.
Here, $\bar y$ is the mean of the output variables in the training data set, and $X$ and $Y$ are matrices of the respective input and output data vectors.


\subsection{Augmented Expression DAG Search Algorithm}

\begin{figure}[ht]
    \centering
    \input{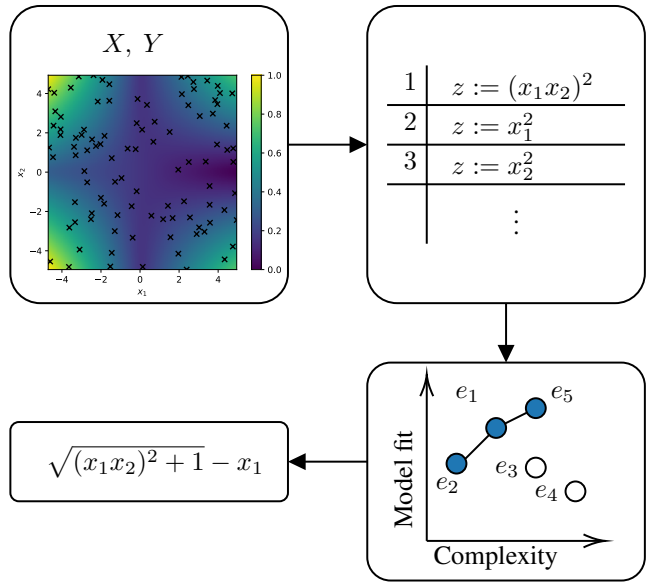}
    \caption{Conceptual sketch of the unbiased, search-based symbolic regression algorithm: From the original problem (1) we select top-$k$ augmentations (2) and solve those problems, creating a Pareto front (3). The returned expression (4) is the best one with complexity below a threshold, or the smallest expression if the complexity of all expressions is above the threshold.}
    \label{fig:workflow}
\end{figure}

We have integrated the variable augmentation into an unbiased, search-based algorithm for symbolic regression that is shown in Figure~\ref{fig:workflow}.
Given a regression problem in terms of a data sample of input and corresponding output variables, our unbiased, search-based symbolic regression algorithm has two main steps:
\begin{enumerate}
\item \textbf{Selecting Variable Augmentations.} Given a standard family $\mathcal{F}$ of regression functions, such as linear regression, polynomial regression, neural networks, and a value $k$, we use the standard family of regression functions and the unbiased expression DAG search to select the top-$k$ scoring variable augmentations.
\item \textbf{Solving Augmented Regression Problems.} For the selected variable augmentations, we compute $k$ symbolic regressors as described in Section~\ref{sec:DAGSearch}. 
For each regressor, we compute its \emph{model} fit in terms of the coefficient of determination $R^2$ and its \emph{model complexity} as the number of nodes in the expression DAG. 
Similar to \cite{feynmanAI2_udrescu20}, we always keep the models on the Pareto front with respect to model fit and model complexity. 
If we have to return a single model, then we return the best fitting model with a complexity below a given threshold. 
If there is no such model, we simply return the smallest model.
\end{enumerate}

\section{Experimental Evaluation}
\label{sec:experiments}

\begin{figure*}[t!]
    \centering
    \begin{tabular}{lcr}
    \includegraphics[width=0.3\textwidth]{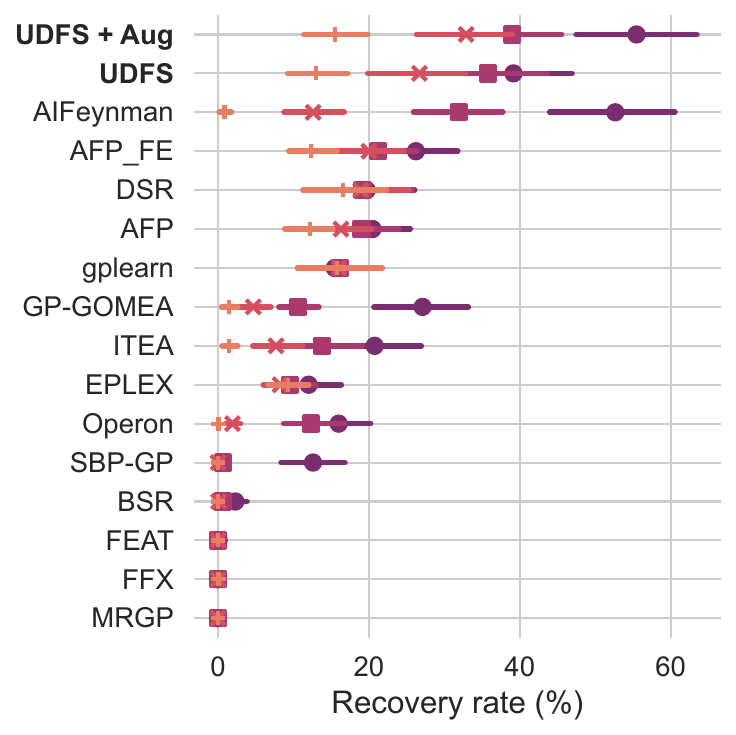} &
    \includegraphics[width=0.3\textwidth]{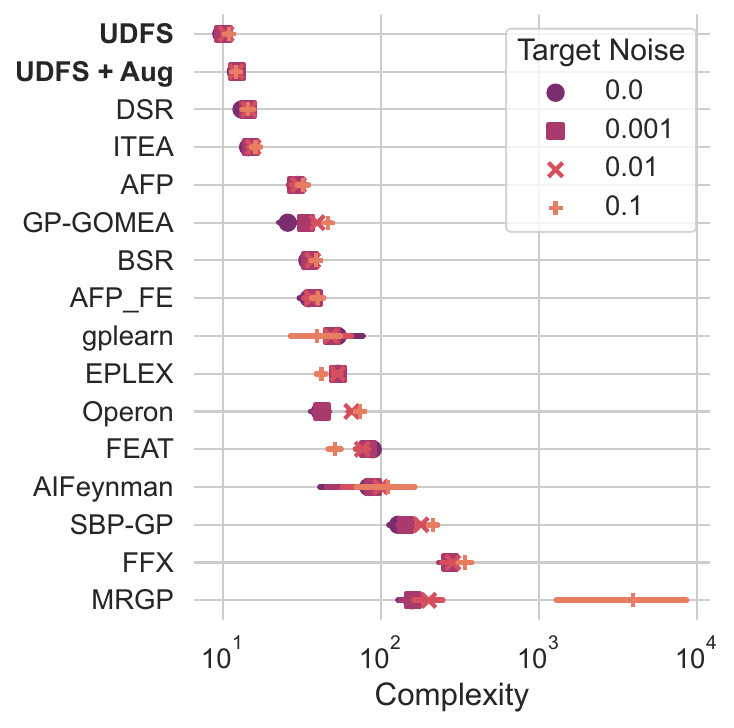} &
    \includegraphics[width=0.3\textwidth]{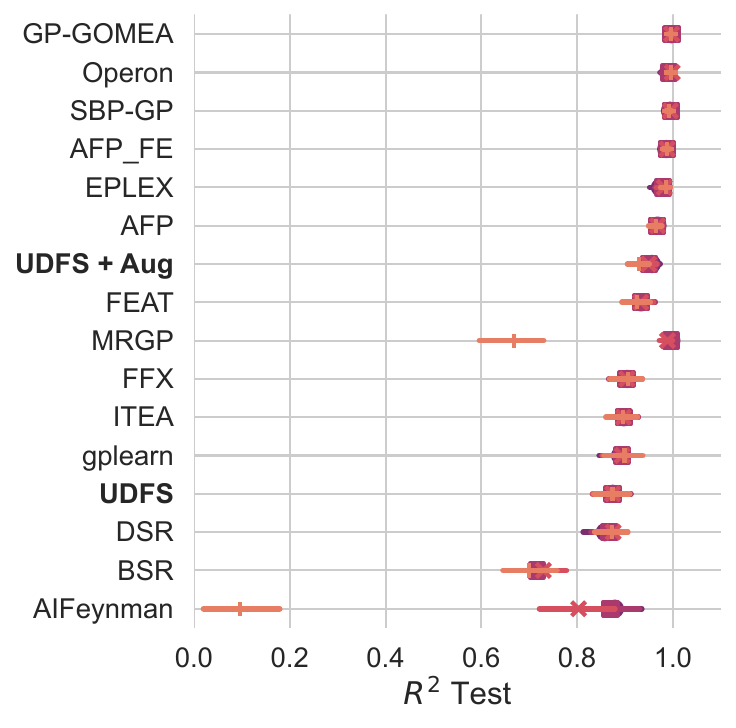}
    \end{tabular}
    \caption{Left: Recovery rate, center: model complexity, right: model fit for \DAG{} and \DAG{} + Aug as scored by the SRBench test suite. The reported values average over 10 runs for the complexity and recovery measures. The reported $R^2$ values are medians from 10 runs, because $R^2$ values can vary significantly.}
    \label{fig:sr_bench_results}
\end{figure*}

We conducted two types of experiments. 
In the first experiment, we compare the performance of our unbiased, search-based approach to the state of the art in symbolic regression on the established and comprehensive SRBench test suite by \cite{srbench_lacava21}. 
In the second experiment, we evaluate the scalability advantage that variable augmentation brings to the unbiased, search-based approach.

Unless stated otherwise, our method, named \DAG{} (Unbiased DAG Frame Search) was used with five intermediary nodes and a maximum of 200\,000 DAG skeletons. 
For the variable augmentation, we have used polynomial regression (\DAG{} + Aug) to select $k=1$ augmentations and up to 30 nodes in the corresponding DAG search.

All experiments were run on a computer with an Intel Xeon Gold 6226R 64-core processor, 128 GB of RAM, and Python 3.10.

\subsection{The SRBench Symbolic Regression Test Suite}

SRBench \cite{srbench_lacava21} is an open-source benchmarking project for symbolic regression.
It comprises 14 of the state-of-the-art symbolic regression models, a set of 252 regression problems from a wide range of applications, and a model evaluation and analysis environment. 
It is designed to easily include new symbolic regression models, such as the \DAG{} + Aug model that we propose here, for benchmarking against the state-of-the-art.  

The models' performance is measured along three dimensions, namely \emph{model fit} on test data, \emph{complexity}, and \emph{accuracy}, that is, the ability to recover ground truth expressions. 
Model fit is measured by the \emph{coefficient of determination} $R^2$, and \emph{complexity} is measured by the number of nodes in the expression tree of an expression. 
Most important for us, a ground truth expression $f$ is considered as \emph{recovered} by a model $\hat{f}$, if either $f - \hat{f}$ can be symbolically resolved to a constant or $\hat{f}$ is non-zero and $\hat{f}/ f$ can be symbolically resolved to a constant.  
The symbolical checks are delegated to the Python library \texttt{SymPy}~\cite{sympy_meurer17}.
The ground truth is only known for 130 out of the 252 problems, namely, the problems from the \emph{Feynman Symbolic Regression Database}\footnote{\url{https://space.mit.edu/home/tegmark/aifeynman.html}} and the \emph{ODE-Strogatz Repository}\footnote{\url{https://github.com/lacava/ode-strogatz}}.
Since we consider ground truth recovery as the most important quality measure for symbolic regressors, as it is a direct measure of interpretability, we restrict ourselves to these 130 problems that are described in the supplemental material.

The comparative experiments summarized in Figure~\ref{fig:sr_bench_results} show that the unbiased search \DAG{} + Aug is on average more \emph{accurate}, that is, it can recover more known ground truth expressions, and also more \emph{robust}, that is, it can recover more ground truth expressions from noisy data, than the state of the art.
Furthermore, \DAG{} + Aug produces small models with an average $R^2$- test score that is close to the optimal score of $1.0$. 
The second-best regressor in terms of recovery, AIFeynman, gives significantly larger and worse fitting models when it cannot recover the ground truth. 
That is, in these cases, AIFeynman is also not a good regressor in the standard sense of regression. 
There is a large group of regression models that produce extremely well fitting models with an average $R^2$ close to $1.0$. 
However, these models are mostly approximations of the ground truth functions, as their recovery rate is significantly lower and the model complexity is quite high.
For a high $R^2$ score, however, one could simply resort to standard regressors, such as neural networks.
But even in terms of standard model fitting performance, \DAG{} and \DAG + Aug place reasonably well among the other symbolic regression methods. 
\begin{figure}[h!]
    \centering
    \includegraphics[width=\columnwidth]{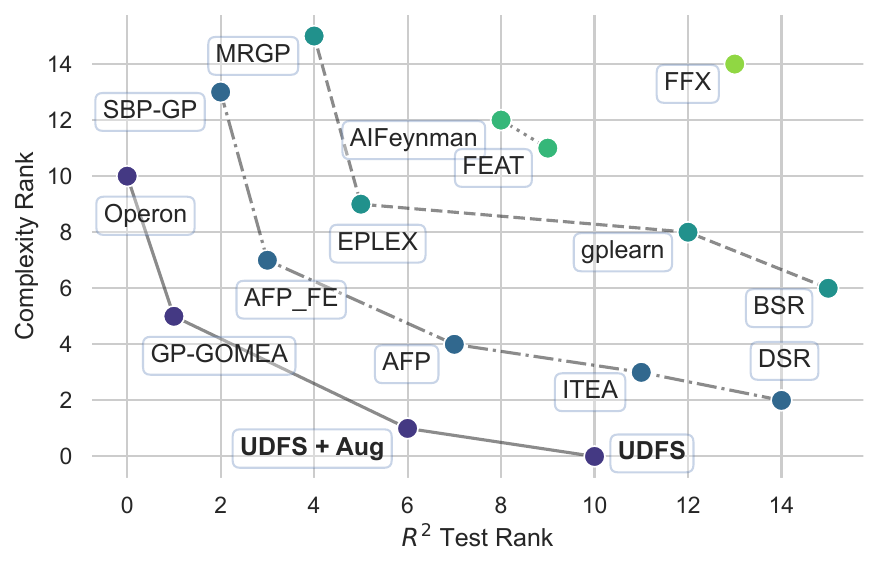}
    \caption{Pareto fronts generated by SRBench for symbolic regression models with respect to the dimensions complexity and model fit ($R^2$ rank).}
    \label{fig:pareto_r2_complexity}
\end{figure}
Figure~\ref{fig:pareto_r2_complexity} shows the different Pareto fronts with respect to the model fit and model complexity. Both, \DAG{} and its variable augmented extension, are on the first Pareto front.

However, both, \DAG{} and \DAG{} + Aug are not perfect.
To gain a better understanding of the expressions that could not be recovered, we did a more fine-grained analysis and also looked at partial recovery of expressions.
For an example, consider the expression \texttt{Feynman\_II\_6\_11} 
\[
\phi(x, y, z) = \cfrac{1}{4\pi\varepsilon_0}\cfrac{p\cos{\theta}}{r^2}
\]
from the Feynman Lectures on Physics~\cite{feynman2011}, where $r,\theta,$ and $p$ are functions of $x,y,$ and $z$, and the permittivity of free space $\varepsilon_0$ is a constant that here is also considered as a regression variable.   
Our search-based regressor returns the expression
\begin{align*}
    \phi(x, y, z) &= \cfrac{1}{10.44\varepsilon_0^{1.5}}\cfrac{p\cos{\theta}}{r^2}\,,
\end{align*}
which fully recovers the functional dependency on the physical variables $r,\theta,$ and $p$, and only misses the correct form of the dependence on $\varepsilon_0$.
That is, the expression returned by the algorithm still provides profound insights into the physics behind the data on which it was run.
Nevertheless, the recovery measure by \cite{srbench_lacava21} considers the problem just as not recovered.
Therefore, we also consider a weaker version of recovery instead.
For comparing two normalized expressions $e_1$ and $e_2$, let $S_1$ and $S_2$ be the corresponding sets of subexpressions.
The Jaccard index~\cite{jaccard_1902}, a similarity measure between sets, 
\[
J(S_1, S_2) = \cfrac{|S_1\cap S_2|}{|S_1\cup S_2|}
\]
can then be used to indirectly measure the similarity of $e_1$ and $e_2$.
We have $J(S_0, S_1)\in [0, 1]$ and $J(S_0, S_1) = 1$, if $e_0$ and $e_1$ have the same subexpressions, which means that they are equal.
For the example given above, we get a Jaccard index of 0.47, reflecting the partial recovery of the ground truth.

\begin{figure}[h!]
    \centering
    \includegraphics[width = 0.67\columnwidth]{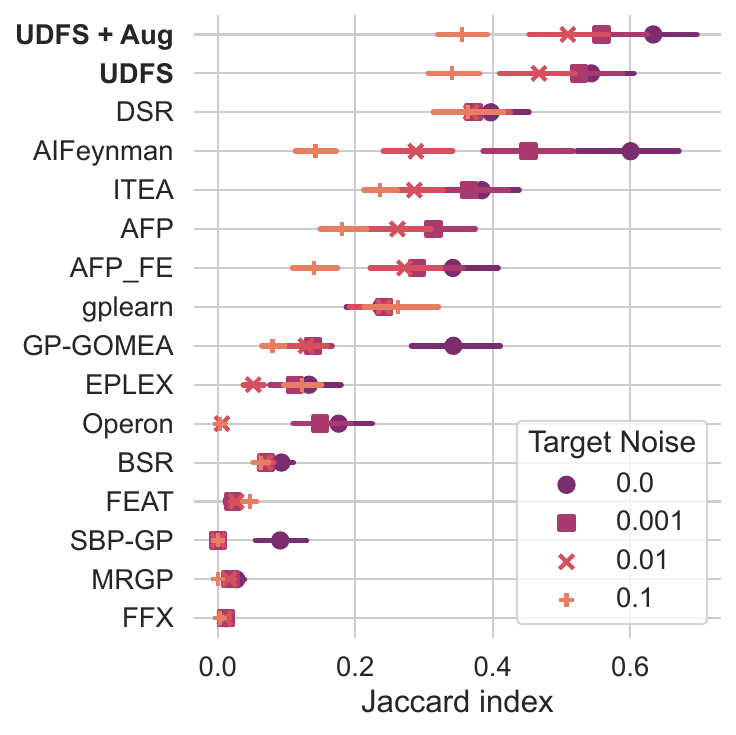}
    \caption{Results of the SRBench test suite for the Jaccard index.}
    \label{fig:jaccard_idx}
\end{figure}

Figure~\ref{fig:jaccard_idx} shows the results of the SRBench test suite using the Jaccard index instead of the full recovery rate. 
Note, that DSR now ranks higher than AIFeynman, because it is more robust against noise. 
The \DAG{} and the \DAG{}+ Aug regressors, however, are still more accurate and robust.

\subsection{Scalability}

\begin{figure}[ht]
    \centering
    \includegraphics[width = \columnwidth]{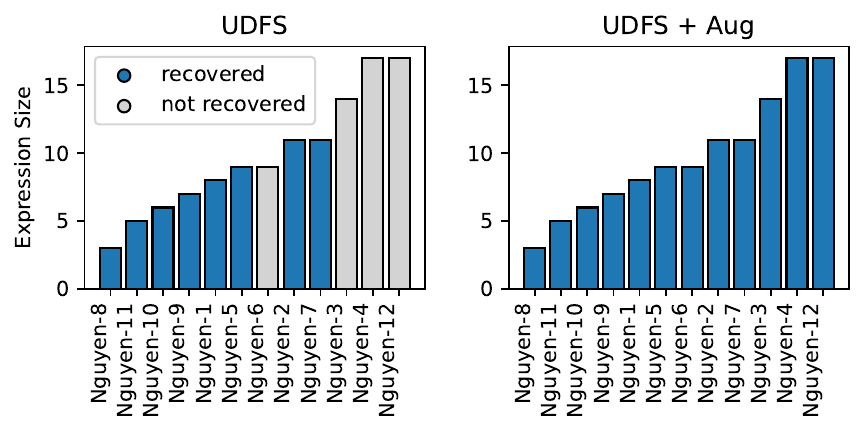}
    \caption{Recovery of Nguyen problems sorted by the number of nodes in the expression tree for the ground truth expression. The shown results are computed with five intermediary nodes and 200\,000 DAG skeletons for \DAG{} (left) and \DAG{} + Aug (right).}
    \label{fig:rec_problem_size}
\end{figure}

The results shown in Figure~\ref{fig:sr_bench_results} show that \DAG{} without variable augmentation compares fairly well to the state of the art. 
\DAG{} + Aug, however, that reduces the effective size of the search space, performs even better.
To see at which search space size variable augmentation does start to improve the performance of \DAG{}, we have compared the performance of \DAG{} and \DAG{} + Aug on the \emph{Nguyen} problems, a collection of small but challenging regression problems that have been used in~\cite{dsr_petersen21,dsr2_petersen21}.
The results are shown in Figure~\ref{fig:rec_problem_size}. 

\begin{figure}[ht]
    \centering
    \includegraphics[width = \columnwidth]{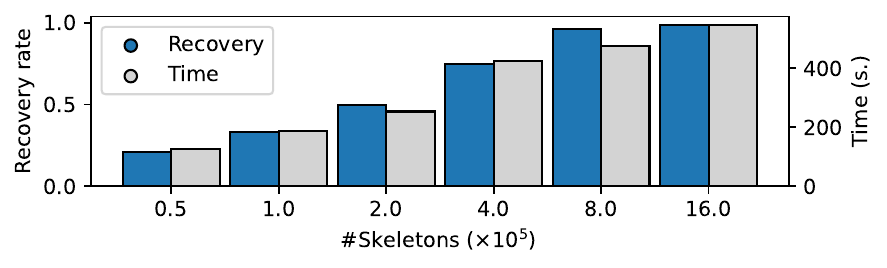}
    \caption{Recovery rate and computation time averaged over 100 runs for the \texttt{Nguyen-6} problem for an increasing number of DAG skeletons with four intermediary nodes.}
    \label{fig:dagframes_nguyen6}
\end{figure}

Interestingly, the \texttt{Nguyen-2} expression $x^4+x^3+x^2+x$, which has a larger expression tree than the \texttt{Nguyen-6} expression $\sin(x) + \sin(x^2+x)$, can be recovered by \DAG, whereas \texttt{Nguyen-6} can not.
\DAG{} however, does not recover the ground truth expression, but the semantically equivalent expression $x^2(x^2 + x) + x^2 + x$, which has a compact DAG representation with only three intermediary nodes that reuse the common subexpressions $x^2$ and $x^2+x$.
The complexity of \DAG{}, however, is not only controlled by the number of intermediary nodes within the DAG frames, but also by the number of DAG skeletons. 
Both, the \texttt{Nguyen-6} and \texttt{Nguyen-7} problems, have DAG representations with four intermediary nodes, but only the latter has been found by \DAG{}. 
The reason why \texttt{Nguyen-6} was not recovered is that the corresponding DAG frame was not included in the search by the random sampling process. 
As can be seen from Figure~\ref{fig:dagframes_nguyen6} the probability to recover \texttt{Nguyen-6} increases with the number of sampled DAG skeletons, but so does the computational effort.

For the \texttt{Nguyen-4} and \texttt{Nguyen-12} problems, that have DAG representations with more intermediary nodes,  it is no longer feasible to increase the number of DAG skeletons to such a degree that they cover the search space reasonably well. 
Both problems, however, benefit from the variable augmentations that are used by \DAG{} + Aug.

More experimental results on scalability, including running times, can be found in the supplemental material.

\section{Conclusions}

Symbolic regression is the problem of searching for a well-fitting function in a space of symbolic function expressions.
Since the search space of potential symbolic function expressions is vast, most symbolic regression approaches are biased in the sense that they make implicit, or sometimes also explicit assumptions, to reduce the effective size of the search space.
Here, we have discussed how to scale up an unbiased search for symbolic regression to problem instances that are used to establish the state of the art in symbolic regression. 
Our unbiased, search-based regressor improves the state of the art in terms of both accuracy and robustness.

\section*{Acknowledgements}
This work was supported by the Carl Zeiss Stiftung within the project "Interactive Inference". In addition, Michael Habeck acknowledges funding by the Carl Zeiss Stiftung within the program "CZS Stiftungsprofessuren".

\bibliographystyle{named}
\bibliography{references}

\clearpage
\onecolumn
\appendix
\noindent
\rule{\textwidth}{0.3pt}
\begin{center}
  \textbf{\LARGE Scaling Up Search-based Symbolic Regression (Supplementary Material)}
\end{center}
\textbf{\hfill submitted to IJCAI 24}
\rule{\textwidth}{0.3pt}

\section{Expression Grammar}
\DAG{} generates mathematical expressions of a context free grammar $G = (N, \Sigma, P, S)$ on $n$-variate functions with at most $p$ constants.
Nonterminals $N$ and terminals $\Sigma$ are defined as
\begin{align*}
    N &= \{S, \text{add},\text{sub},\text{mult}, \text{div}, \text{inv},\text{neg},\text{sine},\text{cosine},\text{logarithm},\text{exponential},\text{sq}, \text{sqrt}\}\\
    \Sigma &= I\cup O\,
\end{align*}
where $I$ and $O$ are inputs and operator symbols
\begin{align*}
    I &= \{x_0,\dots,x_{n-1}, c_0,\dots, c_{p-1}\}\\
    O &= \{+, -, \times, \div, {}^{-1}, \sin, \cos, \log, \exp, {}^2, \sqrt{\phantom{x}}, (, )\}\,.
\end{align*}
The production rules $P$ are then straightforward:
\begin{align*}
    P = \bigcup_{n_1, n_2\in (N\cup I)\setminus\{S\}}\{&S\rightarrow n_1,\\
    &\text{add}\rightarrow (n_1 + n_2),\text{sub}\rightarrow (n_1 - n_2),\text{mult}\rightarrow (n_1 \times n_2),\text{div}\rightarrow (n_1 \div n_2),\\
    &\text{inv}\rightarrow n_1^{-1},\text{neg}\rightarrow -n_1, \text{sine}\rightarrow \sin(n_1),\text{cosine}\rightarrow \cos(n_1),\text{logarithm}\rightarrow \log(n_1),\\
    &\text{exponential}\rightarrow \exp(n_1),\text{sq}\rightarrow n_1^2,\text{sqrt}\rightarrow \sqrt{n_1}\}
\end{align*}
As an example, we could generate an expression for the function $x^2\sin(x^2)$ starting from the start symbol $S$ as follows:
\begin{align*}
    \omit\rlap{\textbf{Expression}}&&&\text{\textbf{Applied Rule}}\\
    S &\rightarrow \text{mult}&&[S\rightarrow n_1]\\
    &\rightarrow (\text{sq}\times\text{sine})&&[\text{mult}\rightarrow (n_1\times n_2)]\\
    &\rightarrow (x^2\times\text{sine})&&[\text{sq}\rightarrow n_1^2]\\
    &\rightarrow (x^2\times\sin(\text{mult}))&&[\text{sine}\rightarrow \sin(n_1)]\\
    &\rightarrow (x^2\times\sin((x\times x)))&&[\text{mult}\rightarrow (n_1\times n_2)]
\end{align*}
The resulting expression $(x^2\times\sin((x\times x)))$ is symbolically equivalent to $x^2\sin(x^2)$.

\section{Sample Space of DAG Skeletons}
\label{sec:sample_space}
Our \DAG{} regressor searches for expressions by exhaustively evaluating every node labeling for a sampled set of DAG skeletons. Given a number of input nodes $n$ and a maximum number of possible intermediary nodes $i$, the sample space is defined by a construction scheme where every node, except the input nodes, chooses its arity and predecessors according to a topological ordering. In the following, we will analyze the size of this sample space based on $n$ and $i$.

\begin{figure}[!ht]
    \centering
     \begin{subfigure}[b]{0.48\textwidth}
         \centering
         \includegraphics[width=\textwidth]{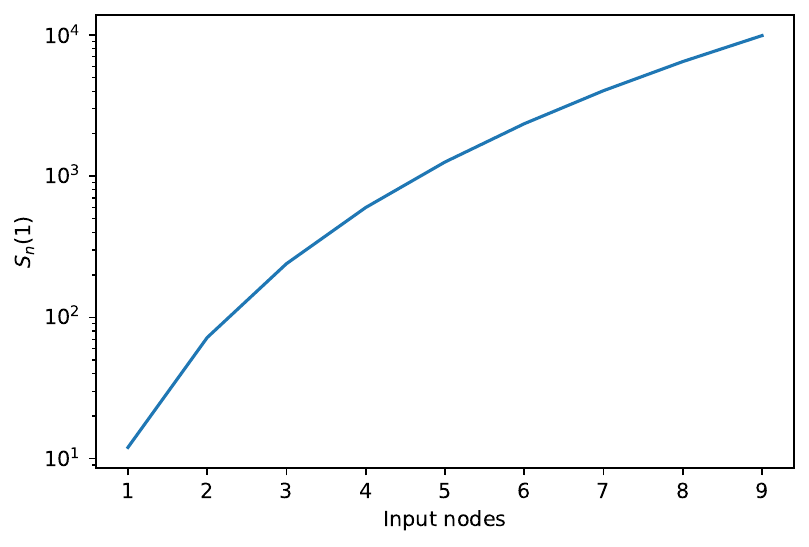}
     \end{subfigure}
     \hfill
     \begin{subfigure}[b]{0.48\textwidth}
         \centering
         \includegraphics[width=\textwidth]{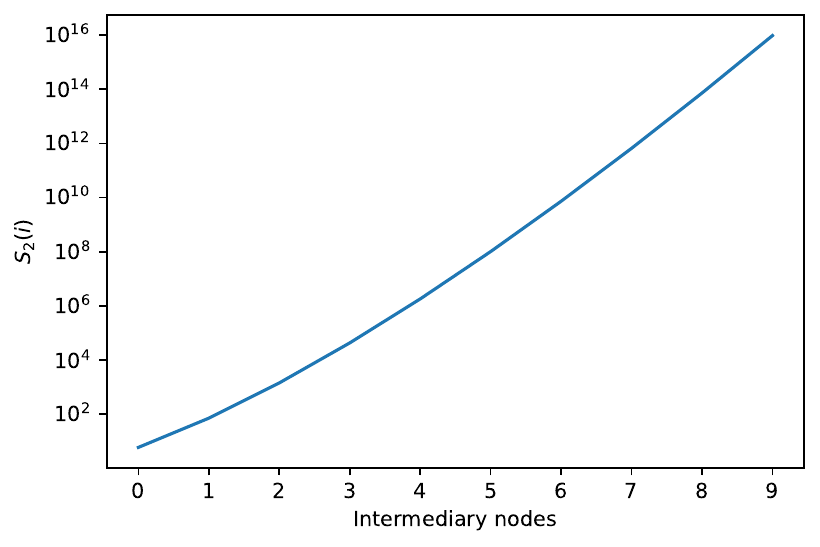}
     \end{subfigure}
    \caption{Size of the sample space for DAG skeletons when increasing the number of input nodes (left) and the number of intermediate nodes (right). The respective fixed dimension was set to $i=1$ and $n=2$.}
    \label{fig:dagframes_recursion}
\end{figure}

During construction, the DAG's output node can select its predecessors from $n + i$ nodes. Hence, there are $(n + i)$ or $(n+i)^2$ possibilities if the output node is unary or binary respectively. So in total, there are  $(n+i) + (n+i)^2$ possibilities for a new output node to select its predecessors.
Adding a new intermediary node can be achieved by treating the current output node as an intermediary node and instead adding a new output node. Each choice of predecessors for this new output node can be combined with every choice of predecessors for the intermediate nodes.
This leads to the following recursion for the number of DAG skeletons $S_n(i)$:
\begin{align*}
    S_n(0) &= n(n+1)\\
    S_n(i) &= S_n(i-1)(n + i)(n + i +1)\,.
\end{align*}
If we unroll this recursion, we get 
\begin{align*}
    S_n(i) &= n(n + i + 1)\prod_{j=1}^i(n+j)^2\,
\end{align*}
which grows exponentially with the number of intermediary nodes $i$ and polynomially with the number of input nodes $n$ as displayed in Figure~\ref{fig:dagframes_recursion}.

This exponential growth makes fully exhaustive search on both, DAG skeletons and node labelings intractable for a larger number of intermediary nodes.

\section{Additions to the Experimental Evaluation Section}

\subsection{Implementation Details}
\label{sec:impl_details}

\subsubsection*{Parameter Optimization}
In order to calculate the score of a DAG $\Delta$ with $p$ parameters, we need to find the best values for the parameters $\theta\in\mathbb{R}^p$ that minimize a loss function $L$:
\begin{align*}
    \textrm{argmin}_{\theta\in\mathbb{R}^p} L(\Delta,\theta).
\end{align*}
One way to optimize the parameters is to pass the DAG to an external non-linear optimizer, such as the Broyden-Fletcher-Goldfarb-Shanno (BFGS) algorithm. However, invoking an external optimizer every time we want to evaluate a DAG is computationally expensive.
\begin{figure}[ht]
    \centering
    \includegraphics[width = 0.5\columnwidth]{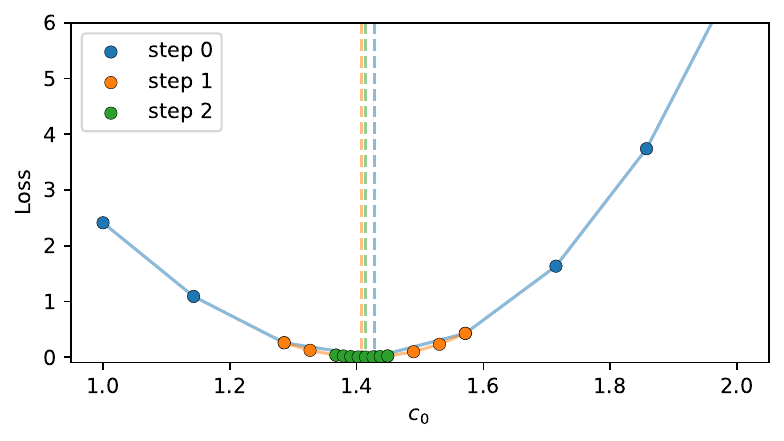}
    \caption{Example for hierarchical grid search for the constant $c$ of the expression $(x - c_0)^2 = (x - \sqrt{2})^2$. Starting with an initial grid (blue) we zoom-in and search in a new grid (orange, green) around the best value (dashed).}
    \label{fig:grid_example}
\end{figure}
Since our implementation allows for a highly parallel evaluation of multiple input variables, we instead use a hierarchical grid search, as depicted in Figure~\ref{fig:grid_example}.
Here we first search in an initial grid. After finding an initial grid value with the best loss, this value is "zoomed in" and searched in a finer grid around this value. This procedure is fast, because we can evaluate a full grid in a single forward pass through the DAG. 
The initial grid defines bounds on the values for $\theta$, since we only focus on values within the grid.
Note however, that this does not restrict the constant values that we can consider in our search, as larger constants can easily be generated by internal operations on the constants $c$ (e.g. $e^{c}$ or $c^2$).
For our implementation, we use grids of size 50 in a two-level hierarchy starting from the interval $[-1, 1]$.

\subsubsection*{Polynomial Models}
Experimental results for \DAG{} + Aug were calculated using augmentations based on polynomial regression.
Here we noticed that in some cases without noise the polynomial regressor was already able to solve the augmented regression problem perfectly.
Therefore we chose to also add the symbolic polynomial model for each augmentation to the set of expressions considered. Hence, if the polynomial symbolic model solves the problem and has a small expression, it will be at the Pareto front and be selected. 

\subsubsection*{Jaccard Index}

For the Jaccard index to be calculated, we need to construct the intersection and union of sets of subexpressions $S_1$ and $S_2$.
For each of those subexpressions, there can be many syntactically different but semantically equivalent expressions.
That is, in order to construct the intersection and the union, one would actually have to compare each subexpression $s_1\in S_1$ to every subexpression $s_2\in S_2$ for semantic equivalence.
A single check for semantic equivalence is very costly. Therefore we omitted the check for semantic equivalence and instead used the string representation of the simplified subexpression according to \texttt{SymPy} by~\cite{sympy_meurer17}.
The resulting sets $S_1$ and $S_2$ are then sets of subexpression strings, for which the intersection and union can be constructed very fast.

\subsection{SRBench}

The SRBench test suite \citep{srbench_lacava21} provides a pipeline to evaluate and compare symbolic regressors on established regression problems.
These regression problems consist of 130 ground truth regression problems and 122 regression problems without a ground truth model. Due to our focus on recovery and the time-consuming evaluation within the provided pipeline, we solely focused on the ground truth problems in this work.
The ground truth problems consist of problems from the \emph{ODE-Strogatz Repository} and \emph{Feynman Symbolic Regression Database} (see Section~\ref{sec:regression_tasks} for more detail).
Results from the SRBench paper for the other symbolic regressors were used as provided by the authors. For the evaluation, each regressor is allowed to provide its own measure of complexity, making it incomparable between the regressors. Hence, we rather used the \texttt{simplified complexity} criterium, which counts the number of nodes in the simplified \texttt{SymPy} expression (including leaves).
To get the pipeline up and running, we used the following workflow, listed below for reproducibility:
\begin{enumerate}
    \item clone the \href{https://github.com/cavalab/srbench}{\textcolor{blue}{SRBench repository}} from \cite{srbench_lacava21} 
    \item add \DAG{} to the \texttt{experiment/methods} directory
    \item run \texttt{experiment/analyze.py} to apply \DAG{} on the ground truth data stored in\\
    \texttt{experiment/pmlb}.
    \item run \texttt{experiment/assess\_symbolic\_model.py} to calculate the recovery rates, $R^2$ scores and simplified complexities. 
    \item run \texttt{postprocessing/collate\_groundtruth\_results.py} to summarize all of the results.
    \item use the \texttt{postprocessing/groundtruth\_results.ipynb} notebook to produce the plots.
\end{enumerate}

\subsection{Robustness}
Results from the SRBench test suite show that a central benefit of our \DAG{} regressor is robustness against noise in the target variable. 
\begin{figure}[!ht]
    \centering
     \begin{subfigure}[b]{0.48\textwidth}
         \centering
         \includegraphics[width=\textwidth]{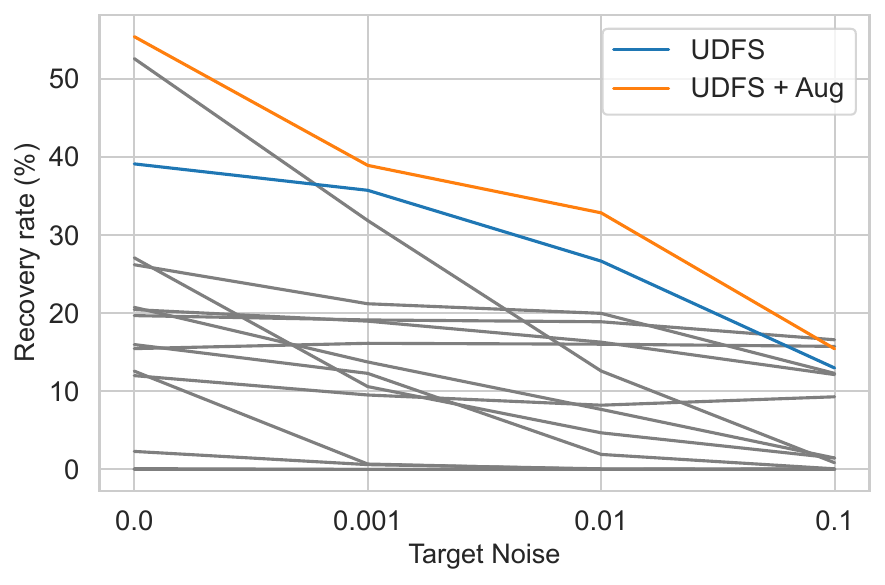}
     \end{subfigure}
     \hfill
     \begin{subfigure}[b]{0.48\textwidth}
         \centering
         \includegraphics[width=\textwidth]{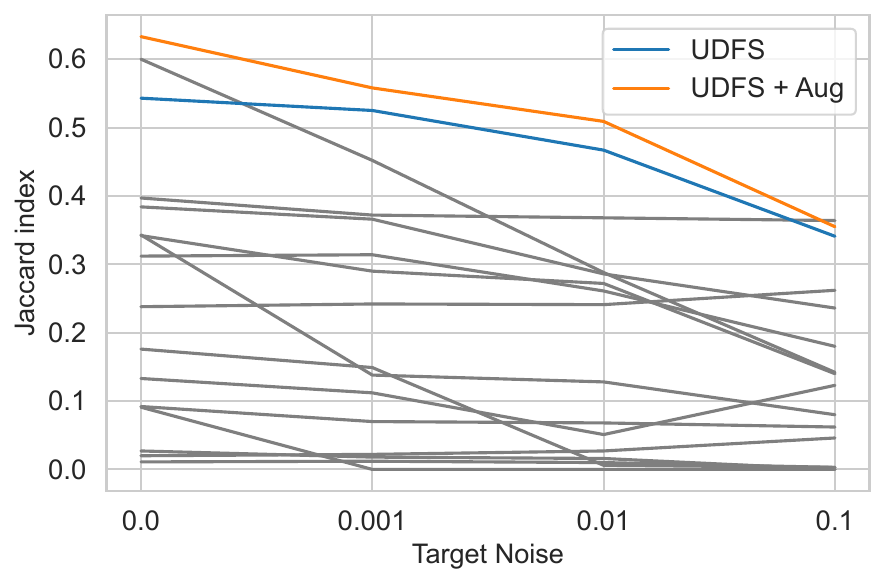}
     \end{subfigure}
     \caption{Average recovery (left) and partial recovery (right) with increasing noise levels. Gray: other regressors used in the SRBench Testsuite.}
     \label{fig:robustness}
\end{figure}
We further illustrate this point in Figure~\ref{fig:robustness}: Our algorithm performs at a very high level and can maintain this high level better than the other regressors for moderate levels of noise. More specifically, at every noise level except the largest, \DAG{} + Aug recovers the most expressions. At the highest noise level, \DAG{} + Aug is still the third best regressor in terms of recovery. 

Another interesting insight is that \DAG{} + Aug dominates \DAG{} completely. With increasing noise, the margin of domination vanishes. This shows that beneficial augmentations can be found at every noise level but noise makes it harder to find them.

\subsection{Scalability}

\subsubsection*{Limits of Scalability}
The complexity of \DAG{} is defined by the number of intermediary nodes $i$ allowed within the DAGs and the maximum number of DAG skeletons $s$ that are considered in the sampling step.
Increasing $i$ will allow more complex expressions, but increase the sample space, whereas a greater $s$ will cover a larger portion of the sample space. Tables~\ref{tab:scalability_feyn} and ~\ref{tab:scaling_recovery_all} show, that scaling up those parameters will increase the performance of \DAG{}.
\begin{table}[ht]
    \centering
    \begin{tabular}{cllll}
    $i$/$s$&10&20&40&80\\
    \toprule
    1&0.11&0.11&0.11&0.11\\
    2&0.33&0.34&0.34&0.34\\
    3&0.37&0.38&0.42&0.42\\
    4&0.39&0.4&0.42&0.43\\
    \end{tabular}
    \caption{Average recovery on Feynman Problems for a different number of intermediary nodes $i$ and a maximum number of DAG skeletons $s$ ($\times 10^3$). Average over 5 tries. Results for the other datasets in Table~\ref{tab:scaling_recovery_all}.}
    \label{tab:scalability_feyn}
\end{table}
\begin{table}[!ht]
    \centering
    \begin{tabular}{cllll||llll||llll}
    $i$/$d$&10&20&40&80&10&20&40&80&10&20&40&80\\
    \toprule
    1&0.2&0.2&0.2&0.2&0.08&0.08&0.08&0.08&0.07&0.07&0.07&0.07\\
    2&0.2&0.2&0.2&0.2&0.17&0.17&0.17&0.17&0.14&0.14&0.14&0.14\\
    3&0.64&0.7&0.7&0.7&0.38&0.38&0.47&0.47&0.19&0.26&0.27&0.29\\
    4&0.7&0.72&0.74&0.72&0.42&0.45&0.53&0.58&0.27&0.27&0.27&0.33\\
    \end{tabular}

    \caption{From left to right: effect on recovery, when increasing the number of intermediary nodes $i$ and the maximum number of DAG skeletons $s$ ($\times 10^3$) for problems of Univ, Nguyen and Strogatz Problems. Average over 5 tries.}
    \label{tab:scaling_recovery_all}
\end{table}

\subsubsection*{Runtime}
\begin{figure}[!ht]
    \centering
    \includegraphics[width = \textwidth]{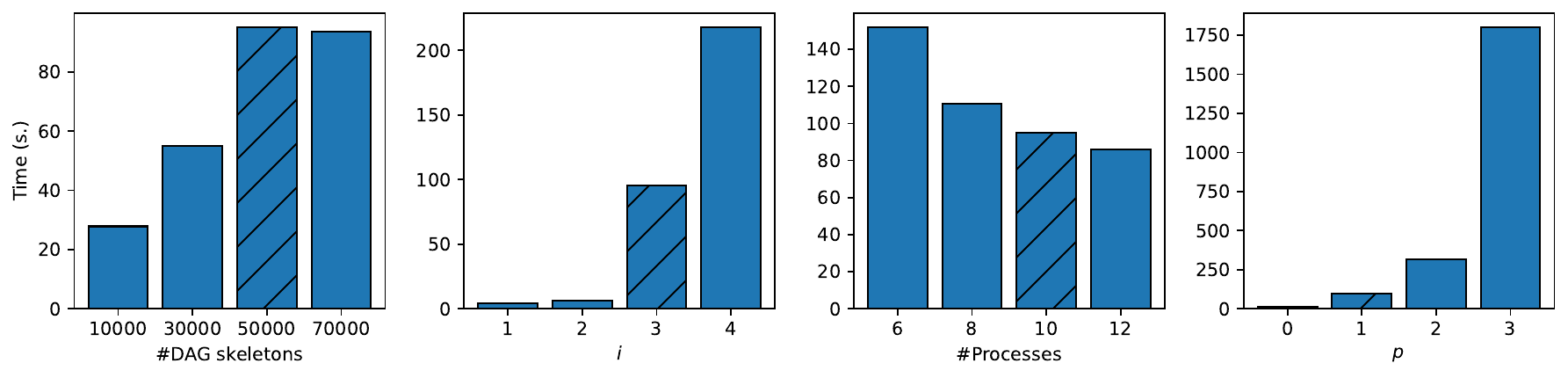}
    \caption{Runtimes when changing a single runtime critical parameter and holding the others fixed at default values (hatched). Average over 10 tries.}
    \label{fig:runtimes}
\end{figure}

In the area of symbolic regression, where researchers are trying to find the "true" expressions underlying the data, computational time can be used more generously than in other areas. However, as the sample space of DAGs grows exponentially with the number of intermediary nodes (see Section~\ref{sec:sample_space}), so does our parallelized computational cost. 
The runtime of \DAG{} is influenced by four parameters: the number of DAG skeletons, intermediary nodes $i$, parameter nodes $p$ and parallel processes. For a default setting of 50\,000 DAG skeletons, three intermediary nodes, one parameter node and 10 parallel processes, we show average runtimes for varying each of those parameters in Figure~\ref{fig:runtimes}. To prevent early stopping, each \DAG{} instance was fitted on a one-dimensional, pure noise regression task $x, y\sim U(0, 1)$.

The results show, that the computational cost grows rapidly with with $i$ and $p$. This is expected as the space of DAG skeletons grows exponentially in $i$ and polynomially in $p$ (see Section~\ref{sec:sample_space}). Additionally, when increasing $p$, we have more effort when optimizing for constants (see Section~\ref{sec:impl_details}).
Increasing the number of DAG skeletons will cover a larger portion of the sample space and leads to a comparatively moderate increase in computational cost. Since for the default settings the search space has size $S_2(3) = 43200$, eventually the number of DAG skeletons becomes larger than the search space.
In this case our implementation becomes fully exhaustive and the computational costs stay the same. Parallelization will decrease execution time, however cannot compensate for the exponential growth when increasing $i$ or $p$. Further steps are therefore necessary to recover larger expressions. For this purpose, we introduced the variable augmentation technique.

\subsubsection*{Augmentation}
\begin{figure}[!ht]
    \centering
     \begin{subfigure}[b]{0.4\textwidth}
         \centering
         \includegraphics[width=\textwidth]{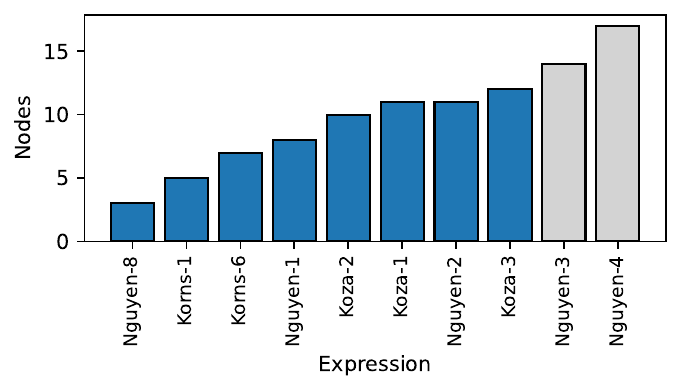}
         \caption{\DAG{}}
     \end{subfigure}
     \hfill
     \begin{subfigure}[b]{0.4\textwidth}
         \centering
         \includegraphics[width=\textwidth]{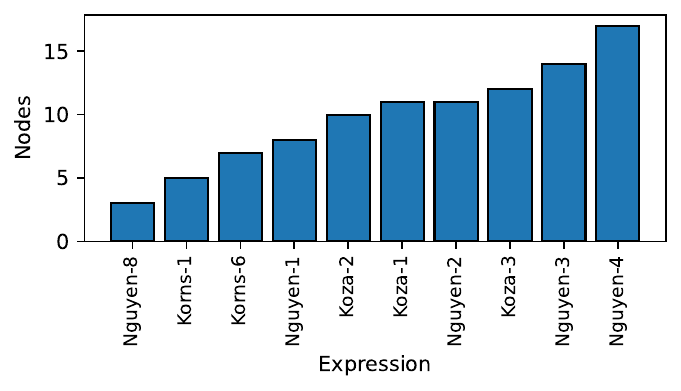}
         \caption{\DAG{} + Aug}
     \end{subfigure}
     \caption{Recovered (blue) and not recovered (gray) \textbf{Univ} Problems.}
     \label{fig:rec_univ}
\end{figure}

\begin{figure}[!ht]
    \centering
     \begin{subfigure}[b]{0.4\textwidth}
         \centering
         \includegraphics[width=\textwidth]{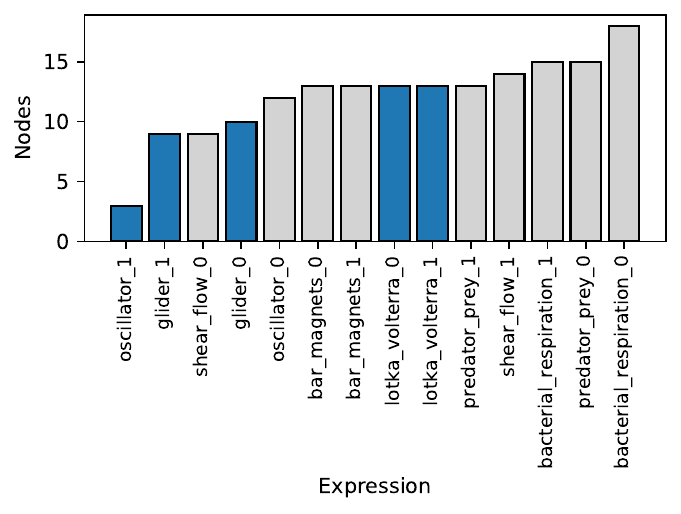}
         \caption{\DAG{}}
     \end{subfigure}
     \hfill
     \begin{subfigure}[b]{0.4\textwidth}
         \centering
         \includegraphics[width=\textwidth]{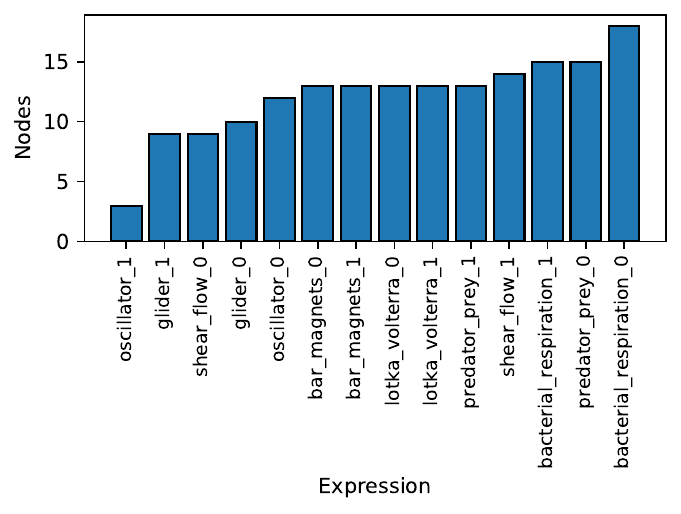}
         \caption{\DAG{} + Aug}
     \end{subfigure}
     \caption{Recovered (blue) and not recovered (gray) \textbf{Strogatz} Problems.}
     \label{fig:rec_strogatz}
\end{figure}

\begin{figure}[!ht]
    \centering
     \begin{subfigure}[b]{\textwidth}
         \centering
         \includegraphics[width=\textwidth]{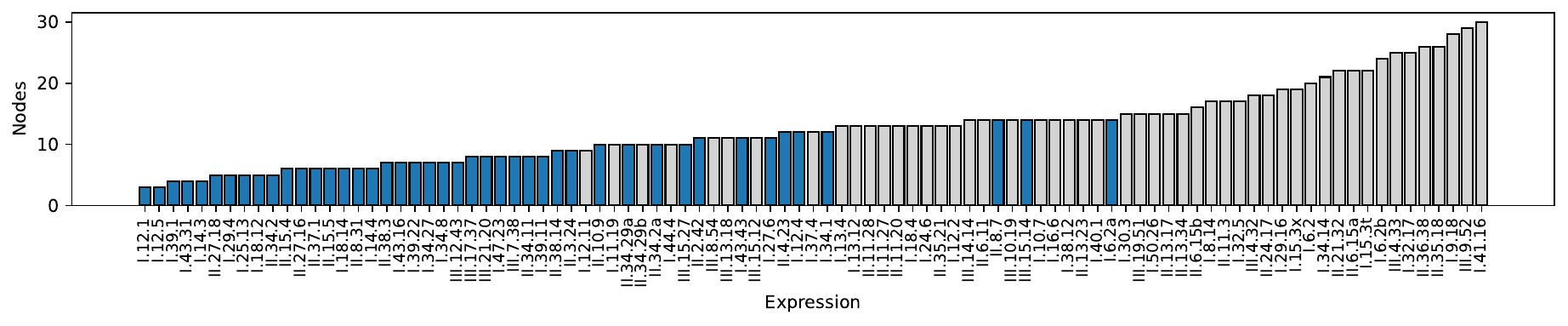}
         \caption{\DAG{}}
     \end{subfigure}
     \hfill
     \begin{subfigure}[b]{\textwidth}
         \centering
         \includegraphics[width=\textwidth]{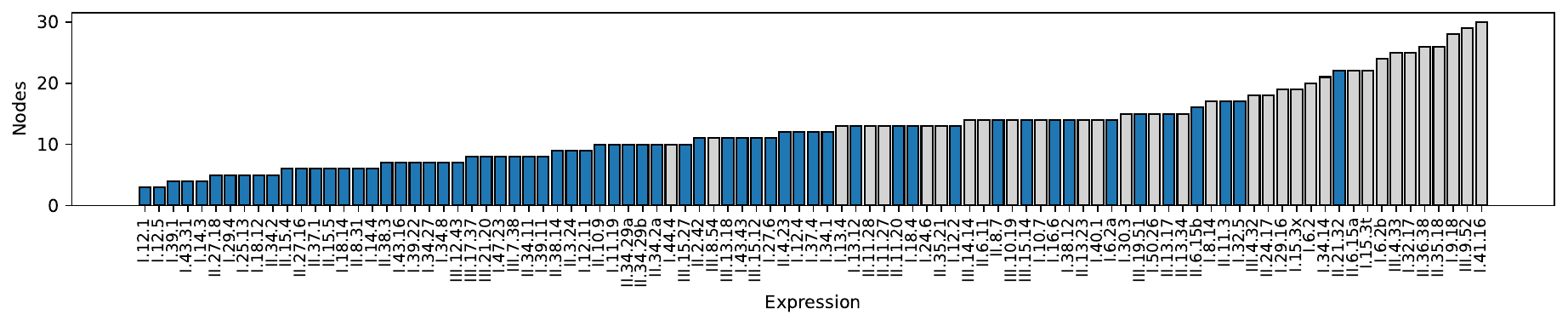}
         \caption{\DAG{} + Aug}
     \end{subfigure}
     \caption{Recovered (blue) and not recovered (gray) \textbf{Feynman} Problems.}
     \label{fig:rec_feynman}
\end{figure}

Variable augmentation allows us to identify beneficial subcomponents in the target expression. Figures~\ref{fig:rec_univ},\ref{fig:rec_strogatz} and \ref{fig:rec_feynman} show, that without augmentation, \DAG{} is able to recover expressions with small expression trees reliably. With augmentation, \DAG{} + Aug can recover significantly more complex expressions.


\subsubsection*{Number of Parameters}
The number of optimizable parameters $p$ in a DAG is itself a hyperparameter. 
We investigated the effect of increasing $p$ in Table~\ref{tab:n_params} for the example expression 
\begin{align*}
    \cfrac{1}{3}x^2\,.
\end{align*}
\begin{table}[ht]
    \centering
    \begin{tabular}{lcccc}
    $p$/$i$&0&1&2&3\\
    \toprule
    0&0.0197&0.0073&0.0002&0.0\\
    1&0.0013&0.0&0.0&0.0\\
    2&0.0013&0.0&0.0&0.0\\
    \end{tabular}
    \begin{tabular}{lcccc}
    $p$/$i$&0&1&2&3\\
    \toprule
    0&False&False&False&True\\
    1&False&True&True&True\\
    2&False&True&True&True\\
    \end{tabular}
    \caption{MSE (left) and Recovery (right) for the expression $\frac{x^2}{3}$ when increasing the number of parameters $p$ for different numbers of intermediary nodes $i$.}
    \label{tab:n_params}
\end{table}
The decreasing MSE for $i=0$ shows that in cases where no recovery is possible, allowing more parameters leads to better fitting models. On the other hand, allowing more when fewer parameters would suffice does not hurt the performance.
More interestingly, with enough intermediary nodes, missing parameters can be compensated. Even when we allow no parameter ($p=0$), \DAG{} was eventually able to recover the true expression via the calculation 
\begin{align*}
    x\div\left(x^{-1} + x^{-1} + x^{-1}\right)\,.
\end{align*}
For our experiments, we chose $p=1$, as a reasonably good tradeoff between computational costs for optimization (see Section~\ref{sec:impl_details}) and possible flexibility of the generated models.

\subsection{Comparison to ESR}

The approach most similar to \DAG{} is the Exhaustive Symbolic Regression (ESR) regressor \cite{esr_bartlett23}. Here, a collection of pre-generated expressions is searched and constants are adjusted to fit a particular regression problem. A notable difference is that ESR can only be used for univariate regression problems. For the following results, we have used ESR with equations up to depth 9 in their expression tree based on the core mathematical operators as defined by \cite{esr_bartlett23}.
\begin{table}[ht]
    \centering
    \begin{tabular}{lcccc}
    &Rec&$R^2$&Complexity&Time\\
    \toprule
    \DAG{} + Aug&1.0&1.0&9.3&26.12\\
    \DAG{}&0.8&1.0&9.4&388.82\\
    ESR&0.5&1.0&9.7&1293.63\\
    \end{tabular}
    \caption{Median $R^2$-Score and average recovery, complexity and execution time (in s.) for \DAG{} and Exhaustive Symbolic Regression (ESR) on the Univ regression tasks.}
    \label{tab:dag_vs_esr}
\end{table}
In Table~\ref{tab:dag_vs_esr} we compare our regressors with the ESR regressor in terms of the evaluation criteria used by SRBench: recovery rate, model fit in terms of the $R^2$ score and complexity as the number of nodes in the expression tree. Additionally, we measured the computation times. While both approaches naturally produce small expressions with good model fit, we can see significant differences in their ability to recover expressions. ESR can only recover 50\% of the Univ tasks, our core algorithm \DAG{} can already recover 80\%. Together with the variable augmentation, \DAG{} + Aug can even solve 100\% of the tasks.

\section{List of Regression Tasks}
\label{sec:regression_tasks}
In the following, we will list the regression problems used in this paper.
\subsection{Nguyen}
The Nguyen problems (Table~\ref{tab:list_nguyen}) have been introduced by \cite{nguyen_problems} in the context of genetic programming. Most recently, \cite{dsr_petersen21,dsr2_petersen21} have used them to compare themselves against the \texttt{Eureqa} framework.
\begin{table}[!ht]
    \centering
    \begin{tabular}{ll}
    Name&Expression\\
    \toprule
    Nguyen-1&$x_{0}^{3} + x_{0}^{2} + x_{0}$\\
    Nguyen-2&$x_{0}^{4} + x_{0}^{3} + x_{0}^{2} + x_{0}$\\
    Nguyen-3&$x_{0}^{5} + x_{0}^{4} + x_{0}^{3} + x_{0}^{2} + x_{0}$\\
    Nguyen-4&$x_{0}^{6} + x_{0}^{5} + x_{0}^{4} + x_{0}^{3} + x_{0}^{2} + x_{0}$\\
    Nguyen-5&$\sin{\left(x_{0}^{2} \right)} \cos{\left(x_{0} \right)} - 1$\\
    Nguyen-6&$\sin{\left(x_{0} \right)} + \sin{\left(x_{0}^{2} + x_{0} \right)}$\\
    Nguyen-7&$\log{\left(x_{0} + 1 \right)} + \log{\left(x_{0}^{2} + 1 \right)}$\\
    Nguyen-8&$x_{0}^{0.5}$\\
    Nguyen-9&$\sin{\left(x_{0} \right)} + \sin{\left(x_{1}^{2} \right)}$\\
    Nguyen-10&$2 \sin{\left(x_{0} \right)} \cos{\left(x_{1} \right)}$\\
    Nguyen-11&$e^{x_{1} \log{\left(x_{0} \right)}}$\\
    Nguyen-12&$x_{0}^{4} - x_{0}^{3} + 0.5 x_{1}^{2} - x_{1}$\\
    \end{tabular}
    \caption{Nguyen Problems}
    \label{tab:list_nguyen}
\end{table}

\subsection{Univ}
The Univ problems (Table~\ref{tab:list_univ}) are a subset of univariate problems collected in \cite{univ_McDermott12} from previous work on genetic programming. The name of each problem is a reference to the authors of the papers in which they used the problem: \texttt{Koza} - \cite{koza_problems}, \texttt{Nguyen} - \cite{nguyen_problems}, \texttt{Korns} - \cite{korns_problems}
\begin{table}[!ht]
    \centering
    \begin{tabular}{ll}
    Name&Expression\\
    \toprule
    Koza-1&$x_{0}^{4} + x_{0}^{3} + x_{0}^{2} + x_{0}$\\
    Koza-2&$x_{0}^{5} - 2 x_{0}^{3} + x_{0}$\\
    Koza-3&$x_{0}^{6} - 2 x_{0}^{4} + x_{0}^{2}$\\
    Nguyen-1&$x_{0}^{3} + x_{0}^{2} + x_{0}$\\
    Nguyen-2&$x_{0}^{4} + x_{0}^{3} + x_{0}^{2} + x_{0}$\\
    Nguyen-3&$x_{0}^{5} + x_{0}^{4} + x_{0}^{3} + x_{0}^{2} + x_{0}$\\
    Nguyen-4&$x_{0}^{6} + x_{0}^{5} + x_{0}^{4} + x_{0}^{3} + x_{0}^{2} + x_{0}$\\
    Nguyen-8&$x_{0}^{0.5}$\\
    Korns-1&$24.3 x_{0} + 1.57$\\
    Korns-6&$0.13 \sqrt{x_{0}} + 1.3$\\
    \end{tabular}
    \caption{Univ Problems}
    \label{tab:list_univ}
\end{table}

\subsection{Strogatz}
The Strogatz problems (Table~\ref{tab:list_strogatz}) are a collection of 2D ordinary differential equations (ODEs). Framed as a regression problem, the goal is to predict the form of the ODE from the given gradients.
The problem collection originates from \cite{strogatz_problems}, the simulated time series and their gradients were taken from the \href{https://github.com/lacava/ode-strogatz}{\textcolor{blue}{repository}} of \cite{strogatz_data}.

\begin{table}[!ht]
    \centering
    \begin{tabular}{ll}
    Name&Expression\\
    \toprule
    bacterial respiration 1&$- \frac{x_{0} x_{1}}{0.5 x_{0}^{2} + 1} - x_{0} + 20$\\
    bacterial respiration 2&$- \frac{x_{0} x_{1}}{0.5 x_{0}^{2} + 1} + 10$\\
    bar magnets 1&$- \sin{\left(x_{0} \right)} + 0.5 \sin{\left(x_{0} - x_{1} \right)}$\\
    bar magnets 2&$- \sin{\left(x_{1} \right)} - 0.5 \sin{\left(x_{0} - x_{1} \right)}$\\
    glider 1&$- 0.05 x_{0}^{2} - \sin{\left(x_{1} \right)}$\\
    glider 2&$x_{0} - \frac{\cos{\left(x_{1} \right)}}{x_{0}}$\\
    lotka volterra 1&$- x_{0}^{2} - 2 x_{0} x_{1} + 3 x_{0}$\\
    lotka volterra 2&$- x_{0} x_{1} - x_{1}^{2} + 2 x_{1}$\\
    predator prey 1&$x_{0} \left(- x_{0} - \frac{x_{1}}{x_{0} + 1} + 4\right)$\\
    predator prey 2&$x_{1} \left(\frac{x_{0}}{x_{0} + 1} - 0.075 x_{1}\right)$\\
    shear flow 1&$\frac{\cos{\left(x_{0} \right)} \cos{\left(x_{1} \right)}}{\sin{\left(x_{1} \right)}}$\\
    shear flow 2&$\left(0.1 \sin^{2}{\left(x_{1} \right)} + \cos^{2}{\left(x_{1} \right)}\right) \sin{\left(x_{0} \right)}$\\
    oscillator 1&$- \frac{10 x_{0}^{3}}{3} + \frac{10 x_{0}}{3} + 10 x_{1}$\\
    oscillator 2&$- \frac{x_{0}}{10}$\\
    \end{tabular}
    \caption{Strogatz Problems}
    \label{tab:list_strogatz}
\end{table}

\subsection{Feynman}
The Feynman problems consist of formulas from the famous Feynman Lectures \cite{feynman2011} and were introduced by \cite{feynmanAI_udrescu20} to test their symbolic regressor. It has since been used and established by \cite{srbench_lacava21} in their comprehensive benchmark suite.
The database can be found \href{https://space.mit.edu/home/tegmark/aifeynman.html}{\textcolor{blue}{here}}.

\begin{longtable}{ll|ll}
    Name&Expression&Name&Expression\\
    \toprule
    \endhead
    \hline
    \multicolumn{4}{c}{Continues on next page}
    \endfoot
    \endlastfoot
    I.10.7&$\frac{x_{0}}{\sqrt{- \frac{x_{1}^{2}}{x_{2}^{2}} + 1}}$&I.11.19&$x_{0} x_{3} + x_{1} x_{4} + x_{2} x_{5}$\\
    I.12.1&$x_{0} x_{1}$&I.12.11&$x_{0} \left(x_{1} + x_{2} x_{3} \sin{\left(x_{4} \right)}\right)$\\
    I.12.2&$\frac{x_{0} x_{1}}{4 \pi x_{2} x_{3}^{2}}$&I.12.4&$\frac{x_{0}}{4 \pi x_{1} x_{2}^{2}}$\\
    I.12.5&$x_{0} x_{1}$&I.13.12&$x_{0} x_{1} x_{4} \cdot \left(\frac{1}{x_{3}} - \frac{1}{x_{2}}\right)$\\
    I.13.4&$\frac{x_{0} \left(x_{1}^{2} + x_{2}^{2} + x_{3}^{2}\right)}{2}$&I.14.3&$x_{0} x_{1} x_{2}$\\
    I.14.4&$\frac{x_{0} x_{1}^{2}}{2}$&I.15.3t&$\frac{- \frac{x_{0} x_{2}}{x_{1}^{2}} + x_{3}}{\sqrt{1 - \frac{x_{2}^{2}}{x_{1}^{2}}}}$\\
    I.15.3x&$\frac{x_{0} - x_{1} x_{3}}{\sqrt{- \frac{x_{1}^{2}}{x_{2}^{2}} + 1}}$&I.16.6&$\frac{x_{1} + x_{2}}{1 + \frac{x_{1} x_{2}}{x_{0}^{2}}}$\\
    I.18.12&$x_{0} x_{1} \sin{\left(x_{2} \right)}$&I.18.14&$x_{0} x_{1} x_{2} \sin{\left(x_{3} \right)}$\\
    I.18.4&$\frac{x_{0} x_{2} + x_{1} x_{3}}{x_{0} + x_{1}}$&I.24.6&$\frac{x_{0} x_{3}^{2} \left(x_{1}^{2} + x_{2}^{2}\right)}{4}$\\
    I.25.13&$\frac{x_{0}}{x_{1}}$&I.27.6&$\frac{1}{\frac{x_{2}}{x_{1}} + \frac{1}{x_{0}}}$\\
    I.29.16&$\sqrt{x_{0}^{2} - 2 x_{0} x_{1} \cos{\left(x_{2} - x_{3} \right)} + x_{1}^{2}}$&I.29.4&$\frac{x_{0}}{x_{1}}$\\
    I.30.3&$\frac{x_{0} \sin^{2}{\left(\frac{x_{1} x_{2}}{2} \right)}}{\sin^{2}{\left(\frac{x_{1}}{2} \right)}}$&I.32.17&$\frac{4 \pi x_{0} x_{1} x_{2}^{2} x_{3}^{2} x_{4}^{4}}{3 \left(x_{4}^{2} - x_{5}^{2}\right)^{2}}$\\
    I.32.5&$\frac{x_{0}^{2} x_{1}^{2}}{6 \pi x_{2} x_{3}^{3}}$&I.34.1&$\frac{x_{2}}{1 - \frac{x_{1}}{x_{0}}}$\\
    I.34.14&$\frac{x_{2} \cdot \left(1 + \frac{x_{1}}{x_{0}}\right)}{\sqrt{1 - \frac{x_{1}^{2}}{x_{0}^{2}}}}$&I.34.27&$\frac{x_{0} x_{1}}{2 \pi}$\\
    I.34.8&$\frac{x_{0} x_{1} x_{2}}{x_{3}}$&I.37.4&$x_{0} + x_{1} + 2 \sqrt{x_{0} x_{1}} \cos{\left(x_{2} \right)}$\\
    I.38.12&$\frac{x_{2}^{2} x_{3}}{\pi x_{0} x_{1}^{2}}$&I.39.1&$\frac{3 x_{0} x_{1}}{2}$\\
    I.39.11&$\frac{x_{1} x_{2}}{x_{0} - 1}$&I.39.22&$\frac{x_{0} x_{1} x_{3}}{x_{2}}$\\
    I.40.1&$x_{0} e^{- \frac{x_{1} x_{2} x_{4}}{x_{3} x_{5}}}$&I.41.16&$\frac{x_{0}^{3} x_{2}}{2 \pi^{3} x_{4}^{2} \left(e^{\frac{x_{0} x_{2}}{2 \pi x_{1} x_{3}}} - 1\right)}$\\
    I.43.16&$\frac{x_{0} x_{1} x_{2}}{x_{3}}$&I.43.31&$x_{0} x_{1} x_{2}$\\
    I.43.43&$\frac{x_{1} x_{3}}{x_{2} \left(x_{0} - 1\right)}$&I.44.4&$x_{0} x_{1} x_{2} \log{\left(\frac{x_{4}}{x_{3}} \right)}$\\
    I.47.23&$\sqrt{\frac{x_{0} x_{1}}{x_{2}}}$&I.50.26&$x_{0} \left(x_{3} \cos^{2}{\left(x_{1} x_{2} \right)} + \cos{\left(x_{1} x_{2} \right)}\right)$\\
    I.6.2&$\frac{\sqrt{2} e^{- \frac{x_{1}^{2}}{2 x_{0}^{2}}}}{2 \sqrt{\pi} x_{0}}$&I.6.2a&$\frac{\sqrt{2} e^{- \frac{x_{0}^{2}}{2}}}{2 \sqrt{\pi}}$\\
    I.6.2b&$\frac{\sqrt{2} e^{- \frac{\left(x_{1} - x_{2}\right)^{2}}{2 x_{0}^{2}}}}{2 \sqrt{\pi} x_{0}}$&I.8.14&$\sqrt{\left(- x_{0} + x_{1}\right)^{2} + \left(- x_{2} + x_{3}\right)^{2}}$\\
    I.9.18&$\frac{x_{0} x_{1} x_{2}}{\left(- x_{3} + x_{4}\right)^{2} + \left(- x_{5} + x_{6}\right)^{2} + \left(- x_{7} + x_{8}\right)^{2}}$&II.10.9&$\frac{x_{0}}{x_{1} \left(x_{2} + 1\right)}$\\
    II.11.20&$\frac{x_{0} x_{1}^{2} x_{2}}{3 x_{3} x_{4}}$&II.11.27&$\frac{x_{0} x_{1} x_{2} x_{3}}{- \frac{x_{0} x_{1}}{3} + 1}$\\
    II.11.28&$\frac{x_{0} x_{1}}{- \frac{x_{0} x_{1}}{3} + 1} + 1$&II.11.3&$\frac{x_{0} x_{1}}{x_{2} \left(x_{3}^{2} - x_{4}^{2}\right)}$\\
    II.13.17&$\frac{x_{2}}{2 \pi x_{0} x_{1}^{2} x_{3}}$&II.13.23&$\frac{x_{0}}{\sqrt{- \frac{x_{1}^{2}}{x_{2}^{2}} + 1}}$\\
    II.13.34&$\frac{x_{0} x_{1}}{\sqrt{- \frac{x_{1}^{2}}{x_{2}^{2}} + 1}}$&II.15.4&$- x_{0} x_{1} \cos{\left(x_{2} \right)}$\\
    II.15.5&$- x_{0} x_{1} \cos{\left(x_{2} \right)}$&II.2.42&$\frac{x_{0} x_{3} \left(- x_{1} + x_{2}\right)}{x_{4}}$\\
    II.21.32&$\frac{x_{0}}{4 \pi x_{1} x_{2} \left(- \frac{x_{3}}{x_{4}} + 1\right)}$&II.24.17&$\sqrt{\frac{x_{0}^{2}}{x_{1}^{2}} - \frac{\pi^{2}}{x_{2}^{2}}}$\\
    II.27.16&$x_{0} x_{1} x_{2}^{2}$&II.27.18&$x_{0} x_{1}^{2}$\\
    II.3.24&$\frac{x_{0}}{4 \pi x_{1}^{2}}$&II.34.11&$\frac{x_{0} x_{1} x_{2}}{2 x_{3}}$\\
    II.34.2&$\frac{x_{0} x_{1} x_{2}}{2}$&II.34.29a&$\frac{x_{0} x_{1}}{4 \pi x_{2}}$\\
    II.34.29b&$\frac{2 \pi x_{0} x_{2} x_{3} x_{4}}{x_{1}}$&II.34.2a&$\frac{x_{0} x_{1}}{2 \pi x_{2}}$\\
    II.35.18&$\frac{x_{0}}{e^{\frac{x_{3} x_{4}}{x_{1} x_{2}}} + e^{- \frac{x_{3} x_{4}}{x_{1} x_{2}}}}$&II.35.21&$x_{0} x_{1} \tanh{\left(\frac{x_{1} x_{2}}{x_{3} x_{4}} \right)}$\\
    II.36.38&$\frac{x_{0} x_{1}}{x_{2} x_{3}} + \frac{x_{0} x_{4} x_{7}}{x_{2} x_{3} x_{5} x_{6}^{2}}$&II.37.1&$x_{0} x_{1} \left(x_{2} + 1\right)$\\
    II.38.14&$\frac{x_{0}}{2 x_{1} + 2}$&II.38.3&$\frac{x_{0} x_{1} x_{3}}{x_{2}}$\\
    II.4.23&$\frac{x_{0}}{4 \pi x_{1} x_{2}}$&II.6.11&$\frac{x_{1} \cos{\left(x_{2} \right)}}{4 \pi x_{0} x_{3}^{2}}$\\
    II.6.15a&$\frac{3 x_{1} x_{5} \sqrt{x_{3}^{2} + x_{4}^{2}}}{4 \pi x_{0} x_{2}^{5}}$&II.6.15b&$\frac{3 x_{1} \sin{\left(x_{2} \right)} \cos{\left(x_{2} \right)}}{4 \pi x_{0} x_{3}^{3}}$\\
    II.8.31&$\frac{x_{0} x_{1}^{2}}{2}$&II.8.7&$\frac{3 x_{0}^{2}}{20 \pi x_{1} x_{2}}$\\
    III.10.19&$x_{0} \sqrt{x_{1}^{2} + x_{2}^{2} + x_{3}^{2}}$&III.12.43&$\frac{x_{0} x_{1}}{2 \pi}$\\
    III.13.18&$\frac{4 \pi x_{0} x_{1}^{2} x_{2}}{x_{3}}$&III.14.14&$x_{0} \left(e^{\frac{x_{1} x_{2}}{x_{3} x_{4}}} - 1\right)$\\
    III.15.12&$2 x_{0} \cdot \left(1 - \cos{\left(x_{1} x_{2} \right)}\right)$&III.15.14&$\frac{x_{0}^{2}}{8 \pi^{2} x_{1} x_{2}^{2}}$\\
    III.15.27&$\frac{2 \pi x_{0}}{x_{1} x_{2}}$&III.17.37&$x_{0} \left(x_{1} \cos{\left(x_{2} \right)} + 1\right)$\\
    III.19.51&$- \frac{x_{0} x_{1}^{4}}{8 x_{2}^{2} x_{3}^{2} x_{4}^{2}}$&III.21.20&$- \frac{x_{0} x_{1} x_{2}}{x_{3}}$\\
    III.4.32&$\frac{1}{e^{\frac{x_{0} x_{1}}{2 \pi x_{2} x_{3}}} - 1}$&III.4.33&$\frac{x_{0} x_{1}}{2 \pi \left(e^{\frac{x_{0} x_{1}}{2 \pi x_{2} x_{3}}} - 1\right)}$\\
    III.7.38&$\frac{4 \pi x_{0} x_{1}}{x_{2}}$&III.8.54&$\sin^{2}{\left(\frac{2 \pi x_{0} x_{1}}{x_{2}} \right)}$\\
    test 1&$\frac{x_{0}^{2} x_{1}^{2} x_{2}^{2} x_{3}^{2} x_{4}^{2}}{16 x_{5}^{2} \sin^{4}{\left(\frac{x_{6}}{2} \right)}}$&test 2&$\frac{x_{0} x_{1} \left(\sqrt{1 + \frac{2 x_{2}^{2} x_{3}}{x_{0} x_{1}^{2}}} \cos{\left(x_{4} - x_{5} \right)} + 1\right)}{x_{2}^{2}}$\\
    test 3&$\frac{x_{0} \cdot \left(1 - x_{1}^{2}\right)}{x_{1} \cos{\left(x_{2} - x_{3} \right)} + 1}$&test 4&$\sqrt{2} \sqrt{\frac{x_{1} - x_{2} - \frac{x_{3}^{2}}{2 x_{0} x_{4}^{2}}}{x_{0}}}$\\
    test 5&$\frac{2 \pi x_{0}^{\frac{3}{2}}}{\sqrt{x_{1} \left(x_{2} + x_{3}\right)}}$&test 6&$\sqrt{\frac{2 x_{0}^{2} x_{1}^{2} x_{6}}{x_{2} x_{3}^{2} x_{4}^{2} x_{5}^{4}} + 1}$\\
    test 7&$\sqrt{\frac{8 \pi x_{0} x_{1}}{3} - \frac{x_{2} x_{3}^{2}}{x_{4}^{2}}}$&test 8&$\frac{x_{0}}{\frac{x_{0} \cdot \left(1 - \cos{\left(x_{3} \right)}\right)}{x_{1} x_{2}^{2}} + 1}$\\
    test 9&$- \frac{32 x_{0}^{4} x_{2}^{2} x_{3}^{2} \left(x_{2} + x_{3}\right)}{5 x_{1}^{5} x_{4}^{5}}$&test 11&$\frac{4 x_{0} \sin^{2}{\left(\frac{x_{1}}{2} \right)} \sin^{2}{\left(\frac{x_{2} x_{3}}{2} \right)}}{x_{1}^{2} \sin^{2}{\left(\frac{x_{2}}{2} \right)}}$\\
    test 12&$\frac{x_{0} \left(- \frac{x_{0} x_{1}^{3} x_{3}}{\left(x_{1}^{2} - x_{3}^{2}\right)^{2}} + 4 \pi x_{2} x_{3} x_{4}\right)}{4 \pi x_{1}^{2} x_{4}}$&test 13&$\frac{x_{0}}{4 \pi x_{4} \sqrt{x_{1}^{2} - 2 x_{1} x_{2} \cos{\left(x_{3} \right)} + x_{2}^{2}}}$\\
    test 14&$x_{0} \left(- x_{2} + \frac{x_{3}^{3} \left(x_{4} - 1\right)}{x_{2}^{2} \left(x_{4} + 2\right)}\right) \cos{\left(x_{1} \right)}$&test 15&$\frac{x_{2} \sqrt{1 - \frac{x_{1}^{2}}{x_{0}^{2}}}}{1 + \frac{x_{1} \cos{\left(x_{3} \right)}}{x_{0}}}$\\
    test 16&$x_{3} x_{5} + \sqrt{x_{0}^{2} x_{1}^{4} + x_{1}^{2} \left(x_{2} - x_{3} x_{4}\right)^{2}}$&test 17&$\frac{x_{0}^{2} x_{1}^{2} x_{4}^{2} \cdot \left(1 + \frac{x_{4} x_{5}}{x_{3}}\right) + x_{2}^{2}}{2 x_{0}}$\\
    test 18&$\frac{3 \left(\frac{x_{1} x_{4}^{2}}{x_{2}^{2}} + x_{3}^{2}\right)}{8 \pi x_{0}}$&test 19&$- \frac{\frac{x_{1} x_{5}^{4}}{x_{2}^{2}} + x_{3}^{2} x_{5}^{2} \cdot \left(1 - 2 x_{4}\right)}{8 \pi x_{0}}$\\

    \caption{Feynman problems}
    \label{tab:list_feynman}
\end{longtable}

\end{document}